\documentclass[11pt,letterpaper]{mystyle}

\usepackage[all]{hypcap}
\usepackage[comma,authoryear,compress]{natbib}
\usepackage{hyperref}[citecolor=lightblue]

\hypersetup{
    colorlinks = true,
    citecolor = {YaleBlue},
}
\usepackage{ulem}
\usepackage{algpseudocode}
\usepackage{subfigure}

\usepackage{microtype}
\usepackage{graphicx}
\expandafter\def\csname ver@subfig.sty\endcsname{}
\usepackage{booktabs} %
\usepackage{float}
\usepackage{bigstrut}
\usepackage{listings}
\usepackage{soul} %

\setstcolor{blue}
\usepackage{amsmath}
\usepackage{amssymb}
\usepackage{mathtools}
\usepackage{amsthm}
\usepackage{mathrsfs}
\usepackage{nicefrac}
\usepackage{dsfont}
\usepackage{enumitem}
\usepackage{subcaption}
\usepackage{graphicx,subfig}
\usepackage{cleveref}
\usepackage{bxcoloremoji}

\usepackage{float}

\usepackage[utf8]{inputenc} %
\usepackage[T1]{fontenc}    %
\usepackage{hyperref}       %
\usepackage{url}            %
\usepackage{booktabs}       %
\usepackage{amsfonts}       %
\usepackage{nicefrac}       %
\usepackage{microtype}      %
\usepackage{graphicx}
\usepackage{subcaption} %
\usepackage{amssymb}
\usepackage{fdsymbol}
\usepackage{wrapfig}
\usepackage{lipsum}
\usepackage{enumitem}
\usepackage{stackengine}
\usepackage[font=small,labelfont=bf]{caption}
\usepackage{color}
\usepackage{adjustbox}
\usepackage{booktabs}    %
\usepackage{tcolorbox}   %
\usepackage{fancyvrb}    %

\usepackage{marvosym}

\usepackage{rotating}
\usepackage{makecell}
\usepackage{xspace}
\usepackage{xcolor}
\usepackage{algorithm}
\tcbuselibrary{breakable}

\newcommand{\x}{\mathbf{x}}

\definecolor{blanchedalmond}{rgb}{1.0, 0.92, 0.8}
\definecolor{carmine}{rgb}{0.59, 0.0, 0.09}
\definecolor{lightblue}{rgb}{0.22,0.45,0.70}%

\renewcommand{\mathbf}{\boldsymbol}

\makeatletter
\def\Ddots{\mathinner{\mkern1mu\raise\p@
\vbox{\kern7\p@\hbox{.}}\mkern2mu
\raise4\p@\hbox{.}\mkern2mu\raise7\p@\hbox{.}\mkern1mu}}
\makeatother

\definecolor{amaranth}{rgb}{0.9, 0.17, 0.31}
\definecolor{antiquebrass}{rgb}{0.8, 0.58, 0.46}
\definecolor{antiquefuchsia}{rgb}{0.57, 0.36, 0.51}
\definecolor{chromeyellow}{rgb}{0.31, 0.47, 0.26}

\newcommand{\github}{\raisebox{-1.5pt}{\includegraphics[height=1.05em]{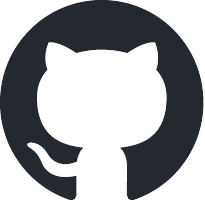}}}

\newtcolorbox{AIbox}[2][]{aibox,title=#2,#1}
\definecolor{lightblue}{rgb}{0.22,0.45,0.70}%
\definecolor{Gray}{gray}{0.95}
\definecolor{Cornsilk}{rgb}{1.0, 0.97, 0.86}
\definecolor{lightorange}{rgb}{0.996, 0.855, 0.643}
\definecolor{lightgray}{rgb}{1.0, 0.827, 0.278}

\usepackage{amsmath}

\usepackage[all]{hypcap}

\newcommand{\methodName}{ReMix\xspace}

\title{Squeeze the Soaked Sponge: \\ Efficient Off-policy Reinforcement Finetuning \\for Large Language Model}

\runningtitle{Squeeze the Soaked Sponge: Efficient Off-policy RFT for Large Language Model}

\author{
  Jing Liang$^1$\textsuperscript{*},
  Hongyao Tang$^1$\textsuperscript{*},
  Yi Ma$^2$,
  Jinyi Liu$^1$,
  Yan Zheng$^1$,
  Shuyue Hu$^3$,
  Lei Bai$^3$,
  Jianye Hao\textsuperscript{\Letter}
}

\affil[1]{College of Intelligence and Computing, Tianjin University}
\affil[2]{School of Computer and Information Technology, Shanxi University}
\affil[3]{Shanghai Artificial Intelligence Laboratory}

\correspondingauthor{Jianye Hao(\href{jianye.hao@tju.edu.cn}{jianye.hao@tju.edu.cn}). \textsuperscript{*}Equal contribution.}

\begin{document}

\begin{abstract}

Reinforcement Learning (RL) has demonstrated its potential to improve the reasoning ability of Large Language Models (LLMs).
Despite the superiority of self-improvement empowered by RL, one major limitation of most existing Reinforcement Finetuning (RFT) methods is that they are \textit{on-policy} RL in nature, i.e., data generated during the past learning process is not fully utilized.
This inevitably comes at a significant cost of compute and time, posing a stringent bottleneck on continuing economic and efficient scaling.
To this end, we launch the renaissance of \textit{off-policy} RL and explore the promise of learning from historical data in the context of RFT. 
Specifically, we propose \textbf{Re}incarnating \textbf{Mix}-policy Proximal Policy Gradient (\textbf{\methodName}), a general approach to enable on-policy RFT methods like PPO and GRPO to leverage off-policy data.
\methodName consists of three major components: 
(1) Mix-policy proximal policy gradient with an increased Update-To-Data (UTD) ratio that utilizes the data generated by both the current policy and past polices for efficient training;
(2) KL-Convex policy constraint that combines the KL constraints on the base model and the precedent model to balance the trade-off between stability and flexibility during training;
(3) Policy reincarnation that replaces the base model with the mix-policy RFT model in the mid way of training and restarts on-policy training, to achieve a seamless transition from efficient early-stage learning to steady asymptotic improvement.
In our experiments, we train a series of \methodName models based on PPO, GRPO from 1.5B, 7B base models.
\methodName achieves an \textbf{average Pass@1 accuracy of 52.10\%} (for 1.5B model) with \textbf{0.079M response rollouts, 350 training steps} and achieves \textbf{63.27\%/64.39\%} (for 7B model) with \textbf{0.007M/0.011M response rollouts, 50/75 training steps} respectively, on five math reasoning benchmarks (i.e., AIME'24, AMC'23, Minerva, OlympiadBench, and MATH500).
Compared with 15 recent advanced models, \methodName shows SOTA-level performance with an over \textbf{30x to 450x reduction in training cost in terms of rollout data volume}, demonstrating superior training efficiency.
In addition, we reveal insightful findings via multifaceted analysis, including 
the implicit preference for shorter responses due to the Whipping Effect of off-policy discrepancy, the collapse mode of self-reflection behavior under the presence of severe off-policyness, the performance under response length constraint, the impact of prompt format, etc.

\vspace{2mm}

\coloremojicode{1F4C5} \textbf{Date}: July 11, 2025 (v3)

\github{} \textbf{Project Page}: \href{https://anitaleungxx.github.io/ReMix}{https://anitaleungxx.github.io/ReMix}

\end{abstract}

\maketitle
\vspace{3mm}

\vspace{-4mm}
\section{Introduction}

\begin{figure}
    \centering
    \includegraphics[width=\linewidth]{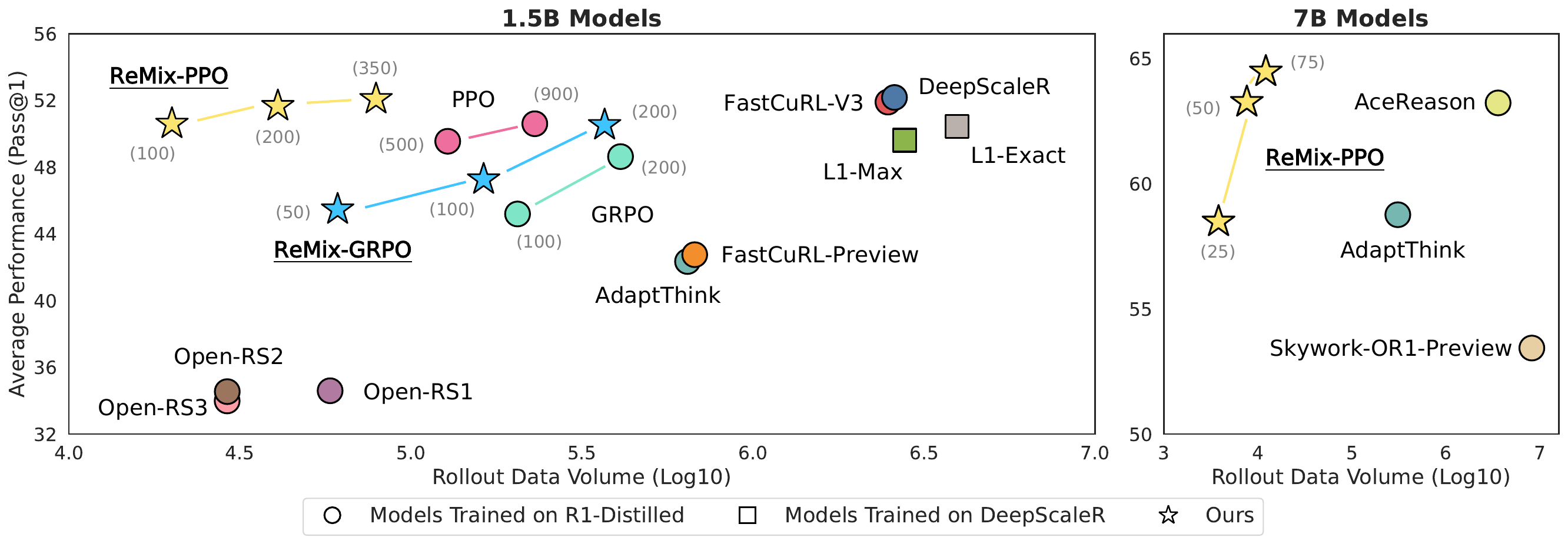}
    \vspace{-0.7cm}
    \caption{\textbf{Efficiency-Performance Comparison for 1.5B Models (\textit{left}) and 7B Models (\textit{right}) in terms of Rollout Data Volume (i.e., total number of responses generated during training) v.s., Average Pass@1 Accuracy on five math reasoning benchmarks.} An ideal model should appear in the top-left corner.
    Our method \underline{\methodName} shows \textbf{superior scores and significantly better training efficiency} compared with standard PPO and GRPO. Moreover, \underline{\methodName-PPO} achieves \textbf{SOTA-level performance} at 1.5B (52.10, 0.079M) and 7B scale (63.27/64.39, 0.007M/0.011M) \textbf{with an over 30x to 450x reduction in rollout data volume} than \texttt{DeepScaleR} (52.14, 2.519M) and \texttt{AceReason-Nemotron} (63.24, 3.584M). The polylines denote the training process with the training step numbers in round brackets.
    }
\label{fig:main}
\end{figure}

The emergence of Large Language Models (LLMs) has lifted artificial intelligence to a next level, with the milestone works like~\citep{openai_gpt3.5,jaech2024openai,bai2022training,trung2024reft,guo2025deepseek}.
Consistent efforts are being made to push forward the limits of LLMs in performing deeper thinking and solving more complex tasks~\citep{li2025system}.
Recently, Large Reasoning Models (LRMs)~\citep{jaech2024openai,guo2025deepseek,team2025kimi,qwen3} have taken the stage and attracted great attention, showing that a significant improvement of problem-solving ability can be achieved by a long human-like reasoning process (i.e., \textit{slow thinking}), especially in scenarios like Math, Coding, Scientific Q\&A, etc.
One of the central recipes of LRMs is Reinforcement Finetuning (RFT)~\citep{trung2024reft}. By treating the LLM as a policy model, the LLM can follow the philosophy of Reinforcement Learning (RL)~\citep{Sutton1988ReinforcementLA} and learn to reason and answer the queries according to the reward signals, e.g., either from a verifiable reward function~\citep{guo2025deepseek} or a learned reward model~\citep{bai2022training}.

Although RFT opens another research space for more powerful reasoning ability beyond Supervised Finetuning (SFT), the longstanding and notorious shortcoming of RL --- \textit{sample inefficiency} --- still exists. In another word, RFT usually needs significantly more computational cost (e.g., autoregressive rollouts via forward inference, gradient updates via network backpropagation) than SFT due to its \textit{trial-and-error} nature in the post-training stage of LLMs.
The inefficiency of RL becomes even more critical in the context of RFT for a larger model and a longer reasoning process.
Consequently, it turns out to be a stringent bottleneck on time and cost that prevents the further scaling of LLMs when pushing the frontier of intelligence.

In the spectrum of RL algorithms, policy gradient algorithms like PPO~\citep{SchulmanWDRK17PPO}, GRPO~\citep{Shao24GRPO}, RLOO~\citep{AhmadianCGFKPUH24RLOO} are widely adopted for RFT of LLMs due to their stable learning performance and friendliness to engineering.
However, all these policy gradient algorithms are \textit{on-policy} algorithms, which are known to be sample inefficient as the data generated by the current policy is dropped after the current iteration of policy training.
In the literature of RL, \textit{off-policy} algorithms are naturally more sample efficient choices since they also learns from the data generated by historical policies during the past learning process (i.e., \textit{experience})~\citep{Sutton1988ReinforcementLA,sutton25experience}.
Following this direction, recent research has begun to incorporate off-policy data in RFT in different ways, including using nonuniform replay strategies~\citep{li2025repo}, learning from positive and negative signals asymmetrically~\citep{roux2025tapered,arnal2025asymmetric}, proposing new learning objectives based on generation consistency~\citep{tang2025rl,cohen2025soft}, and learning from demonstrations of superior models~\citep{yan2025learning}.
Despite the efforts made by these works, off-policy RFT remains underexplored in two aspects:
(1) None of these methods was compared with SOTA on-policy RFT models on multiple mainstream math reasoning benchmarks, leaving training efficiency and final performance of these methods untested thoroughly;
(2) The influence of off-policy learning on the learning process of reasoning ability remains unknown, which impedes essential understanding of off-policy learning for RFT and advancement of effective methodologies.

In this paper, we study off-policy RL for post-training finetuning of LLMs, aiming to achieve SOTA-level reasoning ability efficiently and unbox the effects of off-policy learning for useful insights.
We propose \textbf{Re}incarnating \textbf{Mix}-policy Proximal Policy Optimization (\textbf{\methodName}),
a general approach to enable on-policy proximal policy gradient methods (e.g., PPO and GRPO) to leverage off-policy data efficiently.
\methodName consists of three major components:
(1) Mix-policy proximal policy gradient with an increased Update-To-Data (UTD) ratio~\citep{chenrandomized} leverages the data generated by both current policy and past polices for efficient training;
(2) KL-Convex policy constraint~\citep{ma2024iteratively} combines the KL constraints on the base model and the precedent model to balance the trade-off between stability and flexibility during training;
(3) Policy reincarnation~\citep{AgarwalSCCB22Reincar} replaces the base model with the mix-policy RFT model in the mid way of training and restarts on-policy training, to achieve a seamless transition from efficient early-stage learning to steady asymptotic improvement.
Under the synergy of the three components, \methodName is able to improve the reasoning ability of LLMs efficiently while retaining a stable and flexible training process. 

In our experiments, we adopt PPO and GRPO as representative on-policy methods and implement \textbf{\methodName-PPO} and \textbf{\methodName-GRPO}.
We use \texttt{DeepSeek-R1-Distill-Qwen-1.5B} and \texttt{-7B}~\citep{guo2025deepseek} as the base models, and train our models based on DeepScaleR-Preview-Dataset~\citep{deepscaler2025}.
We conduct a range of comparative evaluations against 15 recent advanced models on five math reasoning benchmarks, including AIME'24, AMC'23, Minerva~\citep{minerva}, OlympiadBench~\citep{olympiadbench}, and MATH500~\citep{math500}.
Figure~\ref{fig:main} summarizes the experimental results in a view of efficiency-performance comparison, for which the detailed discussions are provided in Section~\ref{subsec:main_results}.
Our method achieves an \textbf{average Pass@1 accuracy of 52.10\%} (for 1.5B model) with \textbf{0.079M response rollouts, 350 training steps} and achieves \textbf{63.27\%/64.39\%} (for 7B model) with \textbf{0.007M/0.011M response rollouts, 50/75 training steps} respectively, showing SOTA-level performance and an over \textbf{30x to 450x training cost reduction in terms of rollout data volume}.
Another thing to note is that our models are trained by applying \methodName solely upon PPO and GRPO, leaving possible further improvement from integrating other orthogonal techniques in recent advanced methods in the future.

Moreover, to gain a better understanding of off-policy learning for RFT, we conduct multifaceted studies and analysis, revealing insightful findings including the implicit preference for shorter responses due to the Whipping Effect of off-policy discrepancy, the collapse mode of self-reflection behavior under the presence of severe off-policyness, the performance under response length constraint, the impact of prompt format, etc.

The main contributions of this paper are summarized below:
\vspace{-0.3cm}
\begin{itemize}
    \item We propose \textbf{\methodName}, a general approach designed to enable on-policy proximal policy gradient methods (e.g., PPO and GRPO) to leverage off-policy data generated during the training process for efficient RFT of LLMs.
    
    \item We propose several techniques for controllable utilization of off-policy training that ensure an efficient and stable RFT process while achieving better reasoning performance than on-policy counterparts.
    
    \item We demonstrate the superiority of \methodName in achieving SOTA-level math reasoning ability with a significant reduction in training cost from three aspects (i.e., rollout data volume, training steps, wall-clock time).
    Moreover, we unveil the relationship between off-policy RL and the training dynamics of reasoning behaviors in the context of LLM RFT.

\end{itemize}

\label{sec:intro}
\vspace{-1mm}

\section{Preliminaries}
\label{subsec:preli}

\subsection{Reinforcement Learning for LLM Fine-tuning}

Reinforcement Fine-Tuning (RFT), also referred to as ReFT or RLVR (Reinforcement Learning with Verifiable Rewards), is a paradigm for adapting pre-trained LLMs to specific downstream tasks using RL~\citep{trung2024reft,jaech2024openai}. Unlike methods that rely on human feedback (RLHF), RFT typically employs reward functions that are programmatically determined or based on verifiable outcomes. In this context, \textit{verifiable rewards} mean that the correctness or quality of the LLM's output sequence $\tau$ for a given question $q$ can be assessed automatically (e.g., by comparing to a ground-truth answer, checking against a set of predefined rules, or using an external validation tool), yielding a stationary scalar reward signal.
This paradigm is operationalized by formulating text generation as a sequential decision-making process, specifically a Markov Decision Process (MDP). At each step, the LLM selects a token from its vocabulary to append to the extant sequence. This iterative process continues until a complete output sequence $\tau$ is formed. 

A finite-horizon MDP is formally defined by a tuple $M = (\mathcal{S}, \mathcal{A}, P, R, \gamma)$. 
In the state space $\mathcal{S}$, each \textbf{state $s_t$}~($s_t \in \mathcal{S}$) at timestep $t$ is the inputted question along with the sequence of tokens generated so far, $s_t = (q, y_1, y_2, \dots, y_t)$, where each token $y_i$ is from a finite vocabulary $\mathcal{V}$. The initial state, denoted as $s_0$ sampled from an initial state distribution $\rho_0$, is typically the input prompt or question in a predefined set $\mathcal{D}_0$. Thus an episode commences with the initial prompt $s_0$ and concludes either when the LLM generates a special end-of-sequence token or upon reaching the predefined maximum sequence length $H$. 
The output of such an episode is the generated token sequence $\tau = (y_1, \dots, y_T)$. 
The state space can be defined as $\mathcal{S} = \bigcup_{k=0}^{H} \mathcal{V}^k \times \mathcal{D}_0$. 
An \textbf{action $a_t$}~($a_t \in \mathcal{A}$) involves selecting the next token $y_{t+1}$ from the vocabulary $\mathcal{V}$, where the action space is $\mathcal{A} = \mathcal{V}$. 
The \textbf{transition $P(s_{t+1}|s_t, a_t)$} is deterministic: given a state $s_t = (y_1, \dots, y_t)$ and an action $a_t = y_{t+1}$, the subsequent state becomes $s_{t+1} = (y_1, \dots, y_t, y_{t+1})$. 
Consequently, $P(s_{t+1}|s_t, a_t) = 1$ if $s_{t+1}$ is formed by the concatenation of $s_t$ and $a_t$, and $0$ otherwise.

The \textbf{reward $R(s_t, a_t)$} signal is issued by either a rule-based reward function or a learned reward model usually.
In the scope of this paper, we consider the verifiable reward function. For any non-terminal timestep $t < T-1$, the intermediate reward $R(s_t, a_t)$ is typically $0$. A terminal reward is provided only at the end of the generation process, specifically when the LLM produces the action $a_{T-1}$ that leads to the terminal state $s_T$. For a generated sentence $\tau$, we can denote the reward signal as,
\begin{align}
R(\tau) = 
\begin{cases}
1 & \text{if } \tau \text{ represents the correct answer, with correct format;} \\
0 & \text{otherwise.}
\end{cases}
\end{align}
The \textbf{discount factor $\gamma$}, in light of the sparse terminal reward structure, is often set to $1$. 
The \textbf{policy $\pi_\theta(a_t \mid s_t)$} in the MDP is embodied by the LLM itself, parameterized by $\theta$, and it defines a probability distribution over the selection of the next token $a_t$ given the current sequence of tokens $s_t$.
Following the convention in RL literature, we use $d^{\pi_{\theta}}_{\tau}$ to denote the distribution of the output sequence $\tau$ generated by $\pi_{\theta}$ and use $d^{\pi_{\theta}}_{s,a}, d^{\pi_{\theta}}_{s}$ for the state-action pairs $(s,a)$ and the state respectively. 
The policy performance is denoted by $J(\pi_{\theta}) = \mathbb{E}_{s_0\sim \rho_0, \tau \sim d^{\pi_{\theta}}_{\tau}(s_0)} [R(\tau)]$.
The learning objective of an RL policy is to maximize the reward function regarding the MDP $M$, i.e., $\pi^* = \arg \max_{\pi_{\theta}} J(\pi_{\theta})$. Intuitively, the LLM policy needs to generate correct answers to the target question set $D_0$ to maximize the expectation of reward.

\subsection{Proximal Policy Gradient Methods for Reinforcement Fine-Tuning}

Policy Gradient (PG) methods~\citep{Sutton1988ReinforcementLA} are a mainstream of canonical solutions to the learning objective $J(\pi_{\theta})$ defined in the MDP. To overcome the instability and sample inefficiency of vanilla PG algorithms (e.g., REINFORCE), a series of conservative policy gradient methods~\citep{KakadeL02CPI} and proximal policy gradient (PPG) methods~\citep{SchulmanLAJM15TRPO,SchulmanWDRK17PPO,WangHT19TPPO} have been proposed.
By simplifying the foundational Trust Region Policy Optimization (TRPO)~\citep{SchulmanLAJM15TRPO} algorithm with a clipping mechanism for an easy but effective realization of proximity, Proximal Policy Optimization (PPO)~\citep{SchulmanWDRK17PPO} offers stable training with significantly reduced implementation complexity and computational cost, leading to its widespread adoption. 
In the context of RFT for LLMs, PPO is further developed with a group-based advantage estimator by Group Relative Policy Optimization (GRPO)~\citep{Shao24GRPO}.

We take PPO as a representative of PPG methods here. The primary objective function of PPO, often called the clipped surrogate objective, is given by:
\begin{align}
L^{\text{CLIP}}(\theta) = - \mathbb{E}_{s,a \sim d_{s,a}^{\pi_{\theta_{\text{old}}}}} \Big[ \min\left( r_{\theta}(s, a) \hat{A}(s, a), \text{clip}(r_{\theta}(s, a), 1-\epsilon, 1+\epsilon) \hat{A}(s,a) \right) \Big],
\end{align}
where $r_{\theta}(s, a) = \frac{\pi_\theta(a \mid s)}{\pi_{\theta_{\text{old}}}(a \mid s)}$ represents the importance sampling ratio between the current policy $\pi_\theta$ and the old policy $\pi_{\theta_{\text{old}}}$ (i.e., the policy before the update),
$\hat{A}(s,a)$ is an estimator of the advantage function 
with Generalized Advantage Estimator (GAE)~\citep{SchulmanMLJA15GAE} as a popular choice,
and the clip ratio $\epsilon$ defines the clipping range $[1-\epsilon, 1+\epsilon]$ that determines the proximity of policy update, thereby enhancing stability.
The overall PPO objective function consists of the clipped surrogate objective and a value function loss $L_V(\phi)$ (for training the value function $V_\phi$) and an optional entropy term to encourage exploration. When applying RL for LLM, a KL-divergence penalty is often added to prevent the policy from deviating too far from a reference model $\pi_{\text{base}}$, e.g., the SFT model. The complete objective is:
\begin{align}
L^{\text{PPO}}(\theta, \phi) = \mathbb{E}_{s \sim d_{s,a}^{\pi_{\theta_{\text{old}}}}} \left[ L^{\text{CLIP}}(\theta) - c_1 L_V(\phi) + c_2 \mathcal{H}[\pi_\theta](s) \right] + \beta \cdot \underbrace{\mathbb{E}_{s \sim d^{\pi_{\theta_{\text{old}}}}_{s}} \left[ D_{\text{KL}}(\pi_\theta(\cdot \mid s) || \pi_{\text{base}}(\cdot \mid s)) \right]}_{L_{\text{KL}}(\theta;\pi_{\text{base}})},
\label{eq:typical_ppo_rft}
\end{align}
where $\mathcal{H}[\pi_\theta](s)$ is the entropy of the policy $\pi_\theta$ at state $s$, and $D_{\text{KL}}$ is the KL metric, $c_1$, $c_2$, and $\beta$ are weighting coefficients.
This combined objective is minimized with respect to $\theta$ (for the policy) and $\phi$ (for the value function) concurrently.

\section{Reincarnating Mix-policy Proximal Policy Optimization}
\label{sec:method}

In this section, we introduce our method, Reincarnating Mix-policy Proximal Policy Optimization (\methodName), for efficient and stable RFT of LLMs.
Specifically, \methodName consists of three synergistic innovations: 
(1) Mix-policy proximal policy gradient with an increased Update-To-Data (UTD) ratio for efficient training (Section~\ref{subsec:mixpolicy});
(2) KL-Convex policy constraint to balance the trade-off between stability and flexibility (Section~\ref{subsec:klconvex});
(3) Policy Reincarnation for a smooth transition from efficient early learning to stable asymptotic improvement (Section~\ref{subsec:policyreincarnation}).
We introduce the three components along with the practical implementation (Section~\ref{subsec:imple}) in detail below.

\subsection{Mix-Policy Proximal Policy Gradient with Increased UTD Ratio}
\label{subsec:mixpolicy}

While proximal policy gradient methods like PPO, GRPO deliver strong performance in RFT, the on-policy nature of these methods leads to a significant bottleneck on data utilization. Each optimization iteration necessitates fresh trajectory generation (i.e., on-policy data) through expensive autoregressive forward passes of the policy network, which is a prohibitively costly process for large-scale language models, especially when featured by long reasoning and reflection.

To address this inefficiency, we trace back to the off-policy RL literature for efficient data utilization.
To be specific, we revisit the generalized proximal gradient theory~\citep{QueeneyPC21Geppo}, which extends the \textit{on-policy policy improvement lower bound} originally proposed by~\cite{KakadeL02CPI} and later refined by TRPO~\citep{SchulmanLAJM15TRPO} to a generalized \textit{off-policy policy improvement lower bound}, i.e., Theorem 1 in~\citep{QueeneyPC21Geppo}.
Based on the extension, it allows proximal gradient methods to make use of 
historical trajectories generated during the past policy optimization process while maintaining training stability via temporally constrained importance ratio clipping.

In this work, we launch the renaissance of off-policy RL for efficient RFT and introduce an On-/Off-policy \textit{Mixed} Proximal Policy Gradient method (\textbf{Mix-PPG})
that strategically leverages both off-policy and on-policy data within a unified objective function. 
Formally, for policy at iteration $k$, the mini-batch training data are sampled from a mixture of sources:  
the trajectories generated by historical policies (i.e., $\pi_{k-i}$ for $i \sim \nu$), and the trajectories of the current policy (i.e., $\pi_k$). 
This hybrid sampling strategy balances two competing purposes: (1) Data Reuse: Exploiting past trajectories reduces the autoregressive rollout and inference overhead; (2) Distribution Alignment: Maintaining sufficient on-policy samples prevents training instability and degradation due to the divergence from the current state-action distribution.

\begin{figure}
    \centering
    \includegraphics[width=0.8\linewidth]{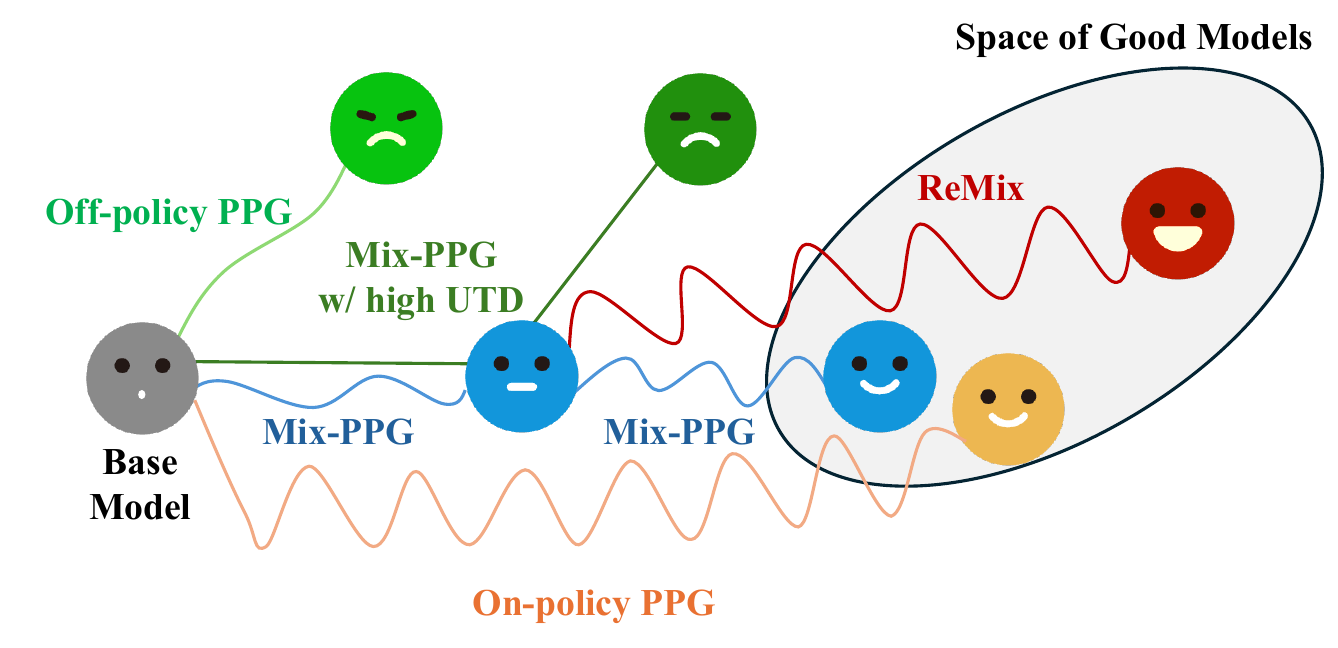}
    \vspace{-0.3cm}
    \caption{\textbf{The conceptual illustration of RFT for LLMs with different proximal policy gradient (PPG) methods (denoted by different colors).} Starting from a base model, (1) the prevalent on-policy PPG methods (e.g., PPO, GRPO) yield a stable and effective training process, yet exhibit inefficient data utilization (i.e., the \textcolor{orange}{orange} waved curve).
    (2) Off-policy PPG offers appealing potential in data efficiency. However, naively adopting off-policy PPG leads to a training collapse (i.e., the less waved \textcolor{green}{green} curve). 
    (3) To strike a balance, we introduce Mix-PPG, which manages to boost early-stage performance but still faces a slow asymptotic improvement (denoted by the \textcolor{cyan}{cyan} curve) and even a collapse when adopting a high UTD ratio (i.e., the straight \textcolor[rgb]{0,0.4,0}{dark green} curve).
    (4) To this end, we propose policy reincarnation and introduce \textbf{ReMix}. ReMix seamlessly takes advantage of both the efficient early-stage training of Mix-PPG and the stable asymptotic improvement of on-policy PPG (i.e., the fusion of the \textcolor{cyan}{cyan} and \textcolor{red}{red} curves), 
    thereby achieving significantly better efficiency at almost no compromise of final performance.}
    \label{fig:policy_reincatenation}
\end{figure}

The policy optimization objective function can be formalized as:
\begin{align}
L_k^{\text{Mix-PPG}}(\theta) = - & \mathbb{E}_{i \sim \nu} \Big[ \mathbb{E}_{(s,a) \sim d^{\pi_{k-i}}_{s,a}} \min \Big( r^{k-i}_{\theta}(s, a) A^{\pi_k}(s,a), \notag\\
& \text{clip} \left( r^{k-i}_{\theta}(s, a), \frac{\pi_k(a \mid s)}{\pi_{k-i}(a \mid s)} - \epsilon, \frac{\pi_k(a \mid s)}{\pi_{k-i}(a \mid s)} + \epsilon \right) A^{\pi_k}(s,a) \Big) \Big],
\label{eq:mix_ppg}
\end{align}
where $i \sim \nu$ with $i \in \{0, 1, \dots, N\}$ is a combined distribution over historical policy indices $\pi_{k-i}$ and the current policy $\pi_k$ (i.e., when $i=0$), the importance sampling ratio $r^{k-i}_{\theta}(s, a) = \frac{\pi_\theta(a \mid s)}{\pi_{k-i}(a \mid s)}$.
Notably, we incorporate a sampling strategy to strike a balance between training stability and efficient data utilization by using a portion $p$ of off-policy data drawn from $\pi_{k-i}$ and $1 - p$ on-policy data drawn from $\pi_k$ with $p \in [0,1]$.
Now, we are ready to replace the on-policy policy optimization objective, e.g., the $L^{\text{Clip}}(\theta)$ term in Eq.~\ref{eq:typical_ppo_rft}, with the Mix-PPG objective $L_k^{\text{Mix-PPG}}(\theta)$ for efficient data utilization.
One thing to note is, we found that explicitly maintaining the portion of on-policy data at a sufficient level is critical to effective training, as much off-policyness will lead to a degradation or even collapse as shown in Figure~\ref{fig:policy_reincatenation}. 
This is why we use the term "Mixed" rather than "Off-policy" for naming our method.
The importance of this nuance is empirically demonstrated later in Figure~\ref{fig:onoffratio} and Figure~\ref{fig:res} of our experiments.

To further improve sample efficiency, we incorporate an increased Update-To-Data (UTD) ratio mechanism, defined originally as the number of gradient updates per environment interaction step. This technique draws inspiration from existing works such as REDQ~\citep{chenrandomized}, which demonstrated that increasing UTD can substantially boost sample efficiency. The core mechanism adopted by us uses a UTD ratio $m$ and performs repeated gradient updates on sampled data batches for $m$ times,
thereby further reducing fresh environment interaction demands. While REDQ pioneers high UTD for value learning in conventional RL problems, we transplant this efficiency to policy optimization in the context of RFT for LLMs.

\subsection{KL-Convex Policy Constraint}
\label{subsec:klconvex}

Conventional RFT imposes a static KL-constraint regularization solely on deviations from the base pre-trained model ${\pi}_{\text{base}}$, inherently suppressing task specialization by over-prioritizing foundational knowledge retention. This rigid static constraint fails to accommodate evolving policy distributions, which could lead to suboptimal updates during the dynamic learning process.

Recent studies~\citep{ma2024iteratively} demonstrate that dual-anchoring to both the base pre-trained model ${\pi}_{\text{base}}$ and the recent historical policy $\pi_{k-1}$ effectively mitigates these issues. Inspired by this, we dynamically update the anchor objective to a convex combination of $\pi_{k-1}$ and ${\pi}_{\text{base}}$. On the one hand, by constraining the policy within the support of ${\pi}_{\text{base}}$, we enforce behavioral consistency with foundational capabilities. This restriction preserves the proximity of the policy distribution to the pre-trained model's generalization properties, thereby preventing catastrophic forgetting of core skills. On the other hand, the constraint imposed on $\pi_{k-1}$ serves as a dynamic adaptation to the policy's current knowledge frontier. It facilitates iterative refinement of the policy based on its previous version $\pi_{k-1}$, thereby enabling the policy to continuously evolve and improve while maintaining a connection to its most recently acquired knowledge. Through this mechanism, the policy can achieve a more robust and comprehensive performance, leveraging the strengths of both the pre-trained model and the iterative refinement process.

Therefore, we reconcile the KL-constraint in RFT via a KL-convex policy constraint (\textbf{KLC}), which modifies the essential optimization objective described in Eq.~\ref{eq:typical_ppo_rft} by replacing the conventional $L_{\text{KL}}(\theta;\pi_{\text{base}})$ with the KL-convex constraint term as formulated below:
\begin{align}
L_{\text{KLC}}(\theta;\pi_{\text{base}}) = \mathbb{E}_{s} \left[ \lambda \cdot D_{\text{KL}}\big( \pi_{\theta}(\cdot \mid s) \parallel {\pi}_{\text{base}}(\cdot \mid s) \big) + (1-\lambda) \cdot D_{\text{KL}}\big( \pi_{\theta}(\cdot \mid s) \parallel \pi_{k-1}(\cdot \mid s) \big) \right],
\label{eq:kl_convex}
\end{align}
where 
$\lambda \in [0,1]$ balances base-model alignment and behavioral consistency with recent policy $\pi_{k-1}$.
This convex combination preserves foundational capabilities while enabling targeted adaptation, acting as a conservative regularizer against over-specialization.

\subsection{Policy Reincatenation}
\label{subsec:policyreincarnation}

While the mix-policy proximal PG method proposed above accelerates early-stage training, the off-policy bias in it can inevitably limit the asymptotic performance, which is widely known in the RL community.
The empirical evidence can be found later in Figure~\ref{fig:res}.
Inspired by Reincarnating RL~\citep{AgarwalSCCB22Reincar}, we adopt the idea and propose Policy Reincarnation in the context of RFT for LLMs.
The purpose of Policy Reincarnation in this paper is to seamlessly combine the advantage of off-policy RL in boosting early-stage training and the stable asymptotic improvement of on-policy RL in the later stage, thus being more efficient at no cost of asymptotic performance.

To be specific, the training process consists of the Mix-PPG stage and the reincarnating on-policy PPG stage.
First, the initial policy model is trained for 
a predetermined $T$ steps of gradient update according to the proposed Mix-PPG algorithm for quick improvement of policy performance.
Thereafter, the reincarnation happens through two changes to the training setting: (1) reset the base model from the initial reference model $\pi_{\text{base}}$ to the current policy model $\pi_{T}$ (which alters the conventional KL constraint term), and (2) switch Mix-PPG to a on-policy PPG method (e.g., PPO or GRPO).

Finally, by composing Mix-PPG (Eq.~\ref{eq:mix_ppg}), KL-convex policy constraint (Eq.~\ref{eq:kl_convex}), and policy reincarnation, we arrive at the complete method proposed in this paper, i.e., Reincarnating Mix-policy Proximal Policy Optimization (\textbf{\methodName}), as follows:
\begin{align}
L^{{\text{ReMix}}}(\theta, \phi) = 
\begin{cases}
\mathbb{E}_{d_{s,a}^{\pi_{\theta}}} \left[ L^{\textcolor{blue}{\text{Mix-PPG}}}(\theta) - c_1 L_V(\phi) + c_2 \mathcal{H}[\pi_\theta](s_t) \right] + \beta \cdot  {L_{\text{KLC}}(\theta;\textcolor{red}{\pi_{\text{base}}})} & \text{if } t \le T \ \text{;}\\
\mathbb{E}_{d_{s,a}^{\pi_{\theta}}} \left[ L^{\textcolor{blue}{\text{PPO}}}(\theta) - c_1 L_V(\phi) + c_2 \mathcal{H}[\pi_\theta](s_t) \right] + \beta \cdot {L_{\text{KLC}}(\theta;\textcolor{red}{\pi_{T}})} & \text{otherwise.}
\end{cases}
\label{eq:remix}
\end{align}
Note that $t$ is the number of batch training steps and the two changes that occur upon policy reincarnation are highlighted in \textcolor{blue}{blue} and \textcolor{red}{red} respectively.
In Eq.~\ref{eq:remix}, we use PPO as the on-policy PPG method for demonstration.
For the case of GRPO, one can remove the value function loss $L_{V}$ and replace the advantage estimation in both $L^{\text{Mix-PPG}}$ and $L^{\text{PPO}}$ with the group-based estimation.
Later in our experiments, we will present the evaluation for both \methodName-PPO and \methodName-GRPO.

The efficacy of ReMix is two-fold.
First, it leverages the advantages of Mix-PPG and on-policy PPG in boosting early-stage training and stable asymptotic improvement respectively, by establishing a seamless transition between the two stages.
Second, the KL-convex policy constraint and the reset of the base reference model for KL constraint (i.e., $\pi_{\text{base}} \rightarrow \pi_{T}$) upon policy reincarnation offers a dynamics and looser constraint compared to the conventional static KL constraint, allowing fast policy training and a larger policy optimization space.
For an intuitive understanding, we provide a conceptual illustration of RFT with different proximal PG methods in Figure~\ref{fig:policy_reincatenation}. The corresponding experimental results can be found later in Figure~\ref{fig:res} of Section~\ref{subsec:analysis}.

\subsection{Practical Implementation}
\label{subsec:imple}

The pseudocode of \methodName is presented in Algorithm~\ref{alg:remix}. 
The training process consists of two stages separated by the policy reincarnation.
For a practical implementation, we need to select the base model and the on-policy proximal PG method first.
We use \texttt{DeepSeek-R1-Distill-Qwen-1.5B} and \texttt{-7B}~\citep{guo2025deepseek} as the specific base models, and we adopt PPO and GRPO as two representative base on-policy proximal PG methods in our experiments.
For the hyperparameters specific to \methodName, we use an off-policy data portion $p=0.4$ for mixed training batch, a UTD ratio $m=2$, a historical policy window size $N=2$, and we set the policy reincarnation step point to $T \in \{50, 100\}$ for \methodName-PPO and $T=50$ for \methodName-GRPO. Especially, the KL-Convex coefficient $\lambda$ decays with training steps $t$ following the rule: \(\lambda(t) = \max(1 - 0.1 \cdot \lceil \max(t - 50,\ 0) / 10 \rceil,\ 0.5)\)
We use the configurations above by default in our experiments, except for the analysis on hyperparameter choice.

\begin{algorithm}[t]
    \small
    \begin{algorithmic}[1]
        \State [Input]: Base model $\pi_{\text{base}}$, and on-policy proximal PG method $\mathbb{A}$ (e.g., PPO, GRPO)
        \State Set training batch size $B$, off-policy data portion $p$, UTD ratio $m$, historical policy window size $N$, policy reincarnation step point $T$
        \State Init the model $\pi_{\theta} = \pi_{\text{base}}$ and the historical policy set $\mathbb{H} = \emptyset$
        \State \textcolor{blue}{\# Stage 1: Mix-policy Proximal PG Training}
        \For{step $t  =  1, 2, 3, ..., T$}
            \State Sample a batch of questions $q \sim \mathcal{D}_0$ with size $(1-p)B$ and generate  fresh responses according to $\pi_{\theta}$ and $\mathbb{A}$
            \State Reuse historical responses from $\mathbb{H}$ with size $pB$ and form the mixed training batch
            \State Save $\pi_{\theta}$ to $\mathbb{H}$ with its responses and logprob data, drop the oldest policy if $|\mathbb{H}|>N$
            \State Repeatedly update $\pi_{\theta}$ with the mixed training batch according to Mix-PPG and $\mathbb{A}$ (the first row, Eq.~\ref{eq:remix}) for $m$ times
        \EndFor
        \State \textcolor{blue}{\# Stage Transition: Policy Reincarnation}
        \State Reset the base reference model from $\pi_{\text{base}}$ to $\pi_{T}$, and drop the historical policy set $\mathbb{H}$
        \State \textcolor{blue}{\# Stage 2: Reincarnating On-policy Proximal PG Training}
        \For{step $t  =  T+1, T+2, T+3, ... $}
            \State Sample a batch of questions $q \sim \mathcal{D}_0$ with size $B$ and generate responses according to $\pi_{\theta}$ and $\mathbb{A}$
            \State Construct a training batch with the fresh responses, and update $\pi_{\theta}$ according to $\mathbb{A}$ (the second row, Eq.~\ref{eq:remix})
        \EndFor
        \end{algorithmic}
    \caption{
    Reincarnating Mix-Policy Proximal Policy Gradient Method (\textbf{\methodName})
    }
\label{alg:remix}
\end{algorithm}

\section{Experiments}
\label{sec:exps}

In this section, we empirically evaluate the efficacy of \methodName on a range of commonly adopted Math reasoning benchmarks, with the detailed experimental setups presented in Section~\ref{subsec:setups}.
We first evaluate the learning performance of \methodName in terms of both the accuracy and the training efficiency against recent LRM baselines (Section~\ref{subsec:main_results}).
Then, we conduct the ablation study to show the contribution of each components of \methodName (Section~\ref{subsec:ablation}).
Moreover, we present the analysis to provide useful insights from different perspectives, including the relationship between off-policy learning and the training dynamics of reasoning behaviors, the performance under constrained maximum response length, the influence of the guide tokens in prompt template, etc. (Section~\ref{subsec:analysis}).

\subsection{Experimental Setup}
\label{subsec:setups}

\paragraph{Training}

We use \texttt{DeepSeek-R1-Distill-Qwen-1.5B} and \texttt{-7B}~\citep{guo2025deepseek} as the base models for our RFT experiments.
As mentioned, our method \methodName is compatible with most on-policy PPG algorithms, therefore we adopt PPO and GRPO as two representative base algorithms in our experiments, resulting in \textbf{\methodName-PPO} and \textbf{\methodName-GRPO}.

We use
DeepScaleR-Preview-Dataset~\citep{deepscaler2025}, which comprises approximately 40,000 unique problem-answer pairs sourced from AIME (1984–2023), AMC (prior to 2023), the Omni-MATH dataset~\citep{gao2024omnimathuniversalolympiadlevel}, and the Still dataset~\citep{stilldataset}.
Following the standard DeepScaler data processing approach, each prompt in the training set was prefixed with "\textless\textbar User\textbar\textgreater " and suffixed with the instruction "Let's think step by step and output the final answer within \text{\textbackslash boxed\{\}}.\ \textless\textbar Assistant\textbar\textgreater\textless think\textgreater" . This structure encourages the model to engage in step-by-step reasoning and produce final answers encapsulated within LaTeX boxed expressions.
One example of the DeepScaler prompt format is shown below. \textcolor{blue}{The blue text} indicates the fixed template used during inference, while the black text represents the instance-specific question inserted into the prompt.

Our experiments are conducted using the \texttt{verl}\footnote{\url{https://github.com/volcengine/verl}} framework and the codebase derived from \texttt{tinyzero}\footnote{\url{https://github.com/Jiayi-Pan/TinyZero}}. 
During training, the model operates with the following generation settings: \texttt{temperature = 1.0}, \texttt{top-p = 1.0}, and \texttt{top-k = -1}. The input prompts are truncated from the right to fit within 766 tokens, and the maximum generation length is \textbf{8,192} tokens.
The detailed hyperparameter choices are presented in Table~\ref{tab:hypers}.

\begin{tcolorbox}[colback=white, colframe=black, title=System Prompt (Standard)]
\textcolor{blue}{\textless\textbar begin\_of\_sentence\textbar\textgreater\textless\textbar User\textbar\textgreater} Xenia and Sergey play the following game. Xenia thinks of a positive integer $N$ not exceeding 5000. Then she fixes 20 distinct positive integers $a_{1}, a_{2}, \ldots, a_{20}$ such that, 
for each $k=1,2, \ldots, 20$, the numbers $N$ and $a_{k}$ are congruent modulo $k$. By a move, Sergey tells Xenia a set $S$ of positive integers not exceeding 20, 
and she tells him back the set $\left\{a_{k}: k \in S\right\}$ without spelling out which number corresponds to which index. How many moves does Sergey need to determine for sure the number Xenia thought of? \textcolor{blue}{Let's think step by step and output the final answer within \text{\textbackslash boxed\{\}}.\ \textless\textbar Assistant\textbar\textgreater\textless think\textgreater}
\end{tcolorbox}

\vspace{-0.5cm}
\paragraph{Evaluation}

We evaluate the performance of different models on a series of mathematical reasoning benchmarks, including AIME'24\footnote{\url{https://huggingface.co/datasets/AI-MO/ aimo-validation-aime}}, AMC'23\footnote{\url{https://huggingface.co/datasets/AI-MO/ aimo-validation-amc}}, Minerva~\citep{minerva}, OlympiadBench~\citep{olympiadbench}, and MATH500\footnote{\url{https://huggingface.co/datasets/HuggingFaceH4/MATH-500}
}. 
Note that all these datasets are not contained in our training dataset, i.e., DeepScaler. 
For the baseline comparisons, we directly download and evaluate the officially released checkpoints from HuggingFace to ensure fair results.
During evaluation, we feed the entire context—including both the prompt and the model-generated response—into the evaluation function. The models in comparison use the same generation settings as in training, except the \texttt{do\_sample} parameter is set to \texttt{false}, resulting in deterministic (greedy) decoding. 
For the evaluation of baseline methods, we use the officially released checkpoints; for our models, we use the best checkpoints obtained within a specific training step budget, e.g., \methodName-PPO (200 Steps).

In our experiments, we focus on the evaluation of our method (i.e., \methodName) in terms of both model performance and training efficiency. 
For the evaluation of model performance, we use  \textbf{Pass@1} accuracy, calculated as the proportion of problems correctly solved on the first attempt.
For training efficiency, we evaluate the models mainly in terms of \textbf{\textit{rollout data volume}}, defined as the total number of rollouts generated by the model during the training process.
It reflects the total amount of model inference, which is usually \textbf{the dominant source} of computational cost during training.
Moreover, we also use \textit{training steps} (i.e., the number of rollout prompt batches) and \textit{training duration} (i.e., the actual elapsed wall-clock time) as additional aspects for efficiency evaluation.

One needs to note that training steps do not reflect the training cost in a fair view, since different models use various batch sizes and rollout settings (e.g., the group size for GRPO).
Therefore, we use rollout data volume as the foundational metric for efficiency evaluation, as it faithfully reflects the cost across different settings and equipments.
Potentially, a more strict efficiency metric should also take the rollout response length into account, however, it varies throughout training and different prompts, making it difficult to accurately calculate. Fortunately, the average response lengths of different models are roughly at the same scale.
Hence, we use rollout data volume.

\vspace{-0.5cm}
\paragraph{Baseline}

We perform comparative evaluations on both 1.5B and 7B scales.
For the comparison among 1.5B models, we evaluate our model (i.e., \texttt{\methodName-R1-Distill-Qwen-1.5B}) against several recent advanced baselines:
\texttt{DeepScaleR-1.5B-Preview}~\citep{deepscaler2025}, \texttt{AdaptThink-1.5B-delta0.1}~\citep{zhang2025adaptthink}, \texttt{FastCuRL-1.5B-Preview}~\citep{song2025fastcurl}, \texttt{II-Thought-1.5B-Preview}~\citep{II-Thought}, \texttt{L1-Qwen-1.5B-Exact}, \texttt{L1-Qwen-1.5B-Max}~\citep{aggarwal2025l1}, \texttt{Open-RS1}, \texttt{Open-RS2}, and \texttt{Open-RS3}~\citep{dang2025reinforcementopenrs}. Except for the L1 series models built on {DeepScaleR-1.5B-Preview}, all others are based on {DeepSeek-R1-Distill-Qwen-1.5B}. 

For the comparison among 7B models, we evaluate our model (i.e., \texttt{\methodName-R1-Distill-Qwen-7B}) against: \texttt{Light-R1-7B-DS}~\citep{lightr1proj}, \texttt{ReasonFlux-F1-7B}~\citep{yang2025reasonflux}, \texttt{Skywork-OR1-7B-Preview}, \texttt{Skywork-OR1-7B}~\citep{he2025skywork}, \texttt{AceReason-Nemotron-7B}~\citep{chen2025acereason}, \texttt{Polaris-7B-Preview} \citep{Polaris2025}, \texttt{AdaptThink-7B-delta0.05}~\citep{zhang2025adaptthink}. All models in this comparison share the same base model, \texttt{DeepSeek-R1-Distill-Qwen-7B}.
We provide a brief overview of the baselines above in Appendix~\ref{app:baseline_intro}.

For existing off-policy RFT methods, we do not include RePO~\citep{li2025repo} because their models are trained under a maximum response length of 1,024 tokens, thus showing limited performance on math reasoning tasks. We do not include LUFFY~\citep{yan2025learning} since the usage of off-policy guidance from a superior model (e.g., DeepSeek-R1) is orthogonal to \methodName, which is also viewed as a different setting where extrinsic guidance or demonstrations are accessible. 
In addition, we did not find public checkpoints for SPO~\citep{cohen2025soft} (which is also trained for code contests), AGRO~\citep{tang2025rl}, AsymRE~\citep{arnal2025asymmetric} and Tapered Off-policy REINFORCE~\citep{roux2025tapered}, thus, we do not include them in our experiments.
Please refer to Section~\ref{sec:related_work} for detailed discussions on related off-policy RFT methods.

\vspace{-0.5cm}
\paragraph{Compute Resource}

The 1.5B model was trained for 50 hours on 2 NVIDIA A800-SXM4-80GB GPUs, while the 7B model required 75 hours on 8 such GPUs. The evaluation of each model was also conducted using the same number of GPUs as in their respective training setups.

For more training details, please refer to Appendix~\ref{app:trainingde}.

\subsection{Performance Evaluation}
\label{subsec:main_results}

The performance evaluation in terms of \text{Pass@1} accuracy on five math reasoning benchmarks are shown in Table~\ref{tab:1.5b} and Table~\ref{tab:7b}, our method \methodName
achieves consistent and substantial improvements over the base 1.5B model~(i.e., \texttt{DeepSeek-R1-Distill-Qwen-1.5B}) and 7B model~(i.e., \texttt{DeepSeek-R1-Distill-Qwen-7B}) across all five benchmarks\footnote{When \texttt{do\_sample} is set to \texttt{true}, Open-RS series models (i.e., -RS1, -RS2, -RS3) show better scores 40.62, 40.08, 39.31 respectively, and \texttt{II-Thought} can achieve a score 51.474. For other models, we found similar scores in our experiments, which do not change the conclusions.}.
Especially for \textbf{\methodName-PPO}, it achieves an \textbf{average performance gain} of \textbf{14.52 points} and \textbf{12.31 points} over 1.5B and 7B base models respectively, \textbf{achieving the second-best average score for 1.5B and the best for 7B among all the baselines and showing the SOTA performance} in the comparison with recent advanced baselines. %
In addition, compared with PPO (900 Steps, 1.5B) and PPO (200 Steps, 7B), our model achieves higher average scores within 100 training steps for 1.5B and 50 steps for 7B, significantly showing the superiority of efficiency.
Similarly, our model exceeds GRPO (100 Steps, 1.5B) and GRPO (200 Steps, 1.5B) within 50 and 200 training steps, respectively.
This indicates that \methodName is able to achieve competitive reasoning ability efficiently with overall no compromise in accuracy and even showing a higher accuracy.

\begin{table}
\centering
\resizebox{\linewidth}{!}{
\begin{tabular}{l|ccccc|cc}
\toprule
\textbf{Model} & \textbf{AIME'24} & \textbf{AMC'23} & \textbf{MATH500} & \textbf{Minerva} & \textbf{Olympiad} & \textbf{Avg.} & \textbf{Cost}\\\midrule
R1-Distilled-Qwen-1.5B (Base) & 33.33 & 43.37 & 67.40 & 16.54 & 27.26 & 37.58 & N/A \\
\midrule
Open-RS1 & 23.33 & 42.17 &  64.20 & 16.18 & 27.11 & 34.60 & \textcolor[rgb]{0,0.6,0}{0.058M} \\
Open-RS2 & 16.67 & 45.78 & 65.00 & 18.38 & 26.96 & 34.56 & \textcolor[rgb]{0,0.6,0}{0.029M}\\
Open-RS3 & 16.67 & 44.58 & 67.60 & 15.64 & 25.48 & 33.99 & \textcolor[rgb]{0,0.6,0}{0.029M}\\
AdaptThink & 13.33 & 57.83 & 78.60 & 23.90 & 38.07 & 42.35 & 0.643M\\
II-Thought & 26.67 & 56.63 & 73.00 & 23.16 & 40.89 & 44.07 & -\\
FASTCuRL-preview & 26.67 & 60.24 & 74.20 & 20.22 & 32.59 & 42.78 & 0.676M\\
FASTCuRL-V3 & 36.67 & 66.27 & \textbf{84.40} & 28.67 & 43.56 & 51.91 & \textcolor{red}{2.478M} \\
L1-Exact* & 23.33 & \textbf{71.08} & \underline{84.00} & \underline{29.41} & \underline{44.59} & 50.48 & \textcolor{red}{3.953M}\\
L1-Max* & 20.00 & \underline{69.88} & 83.00 & 29.04 & \textbf{46.37} & 49.66 & \textcolor{red}{2.764M}\\
DeepScaleR & 40.00 & 65.06 & 83.20 & 29.04 & 43.41 & \textbf{52.14} & \textcolor{red}{2.519M}\\
\midrule
GRPO (100 Steps) & 30.00 &56.63 &75.80 &25.37 &38.22 & 45.20 & 0.205M\\
GRPO (200 Steps) & 36.67 & 61.45 & 80.00 & 25.37 &39.70 &48.64 & 0.410M\\
\rowcolor{Cornsilk} \textbf{\methodName-GRPO (50 Steps)} & 23.33 & 57.83 & 80.40 & 26.10 & 39.70 & 45.47 & \textcolor[rgb]{0,0.6,0}{0.061M} \\
\rowcolor{Cornsilk} \textbf{\methodName-GRPO (100 Steps)} & 23.33 & 62.65 &82.00 & 28.68 &39.70& 47.27& 0.163M \\
\rowcolor{Cornsilk} \textbf{\methodName-GRPO (200 Steps)} & 33.33 & 65.06 & 84.60 & 26.10 & 43.55 & 50.53 & 0.368M\\
\midrule
PPO (500 Steps) & 36.67 & 62.65 & 82.60 & 25.73 & 40.14 & 49.56 & 0.128M\\
PPO (900 Steps) & 30.00 & \underline{69.88} & \underline{84.00} & 25.74 & 43.41 & 50.61 & 0.230M\\
\rowcolor{Cornsilk} \textbf{\methodName-PPO (100 Steps)} & \underline{43.33} & 63.86 &79.60 & 26.84 & 39.41 & 50.61 & \textcolor[rgb]{0,0.6,0}{0.020M} \\
\rowcolor{Cornsilk} \textbf{\methodName-PPO (200 Steps)} & \textbf{46.67} & 62.65 & 82.20 & 26.10 &40.74 & 51.67 & \textcolor[rgb]{0,0.6,0}{0.041M} \\
\rowcolor{Cornsilk} \textbf{\methodName-PPO (350 Steps)} & 36.67\textsuperscript{\textcolor{red}{↑3.34}} & \underline{69.88} \textsuperscript{\textcolor{red}{↑26.51}} & 82.00\textsuperscript{\textcolor{red}{↑14.60}} & \textbf{30.15}\textsuperscript{\textcolor{red}{↑13.61}} & 41.78\textsuperscript{\textcolor{red}{↑14.52}} & \underline{52.10}\textsuperscript{\textcolor{red}{↑14.52}} & \textcolor[rgb]{0,0.6,0}{0.079M} \\
\bottomrule
\end{tabular}}
\vspace{-0.2cm}
\caption{\textbf{Pass@1 Accuracy (\%) and Training Cost (in terms of Rollout Data Volume) of 1.5B Models.}
The maximum response generation length is 8,192 tokens.
\textbf{Bolded} and \underline{underlined} values denote the highest and the second-highest scores in each dataset (i.e., column).
\textbf{\methodName achieves better average scores than both the standard PPO and GRPO in a significantly more efficient manner}, and \textbf{\methodName-PPO achieves the second-best average score} across five math reasoning benchmarks \textbf{within 350 training steps}. 
The arrow \textcolor{red}{↑} denotes the improvement over the base model.
For cost, we mark the costs \textcolor{red}{$> 1$M} and \textcolor[rgb]{0,0.6,0}{$< 0.1$M} in corresponding colors, `-' denotes that not enough information was found. 
}
\label{tab:1.5b}
\vspace{-0.2cm}
\end{table}

\begin{table}
\centering
\resizebox{\linewidth}{!}{
\begin{tabular}{l|ccccc|cc}
\toprule
\textbf{Model} & \textbf{AIME'24}  & \textbf{AMC'23} & \textbf{MATH500} & \textbf{Minerva} & \textbf{Olympiad} & \textbf{Avg.} & \textbf{Cost}\\\midrule
R1-Distilled-Qwen-7B (Base) & 33.33 & 68.68 & 83.80 & 30.15 & 44.44 & 52.08 & N/A \\
\midrule
ReasonFlux-F1 & 20.00 & 54.22 & 77.20 & 29.04 & 37.04 & 43.50 & - \\
Light-R1 & 30.00 & 66.27 &87.00 & 34.56 & 47.56 &53.08 & - \\
Skywork-OR1-Preview & 43.33 & 63.86 & 84.40 &29.41 & 46.22 & 53.44 & \textcolor{red}{>8.192M}\\
Polaris & 40.00 & 63.86 &87.60 & 36.40 &48.00 &55.17 & -\\
AdaptThink &  46.67 &  75.90 & 87.60 & 33.46 &  50.22 &  58.77 & 0.307M\\
AceReason-Nemotron & \underline{60.00} & \textbf{80.72} & {89.00} & 36.40 & 50.07 & {63.24} & \textcolor{red}{>3.584M}\\
\midrule
PPO (50 Steps) &  33.33 & 71.08 & 87.20 & 36.03 & 48.00 & 55.13 & \textcolor[rgb]{0,0.6,0}{0.013M}\\
PPO (100 Steps) &  40.00 &77.11 & \underline{90.00} & 35.66 & \underline{51.56} &58.87 & \textcolor[rgb]{0,0.6,0}{0.026M}\\
PPO (200 Steps) & 53.33 &78.31 &87.00 &34.19 & 48.88 & 60.34 & \textcolor[rgb]{0,0.6,0}{0.051M}\\
\rowcolor{Cornsilk} \textbf{\methodName-PPO (25 Steps)}& 36.67 & 78.31 & {89.00} & \underline{38.24} &50.22 &58.49 & \textcolor[rgb]{0,0.6,0}{0.003M}\\
\rowcolor{Cornsilk} \textbf{\methodName-PPO (50 Steps)} & 56.66 & \underline{79.52} & 88.60  & \textbf{38.97} & \textbf{52.59} &\underline{63.27} & \textcolor[rgb]{0,0.6,0}{0.007M} \\
\rowcolor{Cornsilk} \textbf{\methodName-PPO (75 Steps)}& \textbf{63.33}\textsuperscript{\textcolor{red}{↑30.00}} & 78.31\textsuperscript{\textcolor{red}{↑9.63}} & \textbf{90.20}\textsuperscript{\textcolor{red}{↑6.40}} & {37.50}\textsuperscript{\textcolor{red}{↑7.35}} & \textbf{52.59}\textsuperscript{\textcolor{red}{↑8.15}} &\textbf{64.39}\textsuperscript{\textcolor{red}{↑12.31}}  & \textcolor[rgb]{0,0.6,0}{0.011M}\\
\bottomrule
\end{tabular}}
\vspace{-0.2cm}
\caption{\textbf{Pass@1 Accuracy (\%) and Training Cost (in terms of Rollout Data Volume) of 7B Models.}
The maximum response generation length is 8,192 tokens.
\textbf{Bolded} and \underline{underlined} values denote the highest and the second-highest scores in each dataset (i.e., column).
\textbf{\methodName-PPO achieves the best average score} across five math reasoning benchmarks \textbf{within 75 training steps}. We did not find better checkpoints for PPO within 500 steps, and for \methodName-PPO within 200 steps. \methodName-GRPO is not included for 7B-scale comparison due to computational resource constraints. The arrow \textcolor{red}{↑} denotes the improvement over the base model.
`-' denotes that not enough information was found. 
}
\label{tab:7b}
\end{table}

More importantly, we move on to the evaluation in terms of training efficiency.
This is shown in the last volume (i.e., Cost)of Table~\ref{tab:1.5b} and~\ref{tab:7b}, and notably, Figure~\ref{fig:main} illustrates \textbf{the efficiency–accuracy trade-off} of 1.5B models and 7B models in terms of \textbf{rollout data volume} (\textbf{log$_{10}$ scale}) versus \textbf{average Pass@1 accuracy}, where the scores are out of Table~\ref{tab:1.5b} and~\ref{tab:7b} (i.e., \textbf{Avg.} and \textbf{Cost}).
In the ideal case, the model should appear in the top-left corner of the plot, and therefore the closer a model is to the top-left corner, the better the model is. 
To ensure a fair comparison, the rollout data volume of square-marked models (which means the models fine-tuned upon \texttt{DeepScaleR}) includes the data cost of training \texttt{DeepScaleR} itself. 
For \methodName-GRPO and GRPO, we report results after 200 training steps due to computational resource constraints.
For clarity, we summarize the major observations in Figure~\ref{fig:main} below:
\vspace{-0.4cm}
\begin{itemize}
    \item \textbf{(1.5B) ReMix-PPO v.s., DeepScaleR:}
    DeepScaleR, the strongest 1.5B competitor, requires around \textbf{2.519M} rollouts to reach its final score (i.e., \textbf{52.14}), whereas \methodName-PPO (350 Steps) achieves a comparable score (i.e., \textbf{52.10}) with \textbf{0.079M} rollouts --- over \textbf{a 30x reduction} in rollout data volume.
    
    \item \textbf{(1.5B) ReMix-PPO v.s., PPO:} We trace the performance of \methodName-PPO at 100, 200, and 350 training steps (denoted by the \textbf{\textcolor[rgb]{0.85,0.85,0}{yellow curve}} in Figure~\ref{fig:main}), corresponding to rollout data volumes of roughly 0.020M, 0.041M, and 0.079M, respectively.
    Even after generating just \textbf{0.020M} rollout samples, \methodName-PPO achieves a score of \textbf{50.61}, which has already surpasses most baselines. Compared to PPO (900 Steps), which achieves an average score of \textbf{50.61} with \textbf{0.230M} rollouts, our model shows over \textbf{a 10x reduction} in rollout data volume.

    \item \textbf{(1.5B) ReMix-GRPO v.s., GRPO:} We also trace the performance of \methodName-GRPO at 50, 100, and 200 training steps (denoted by the \textbf{\textcolor{cyan}{cyan curve}} in Figure~\ref{fig:main}), corresponding to rollout data volumes of roughly 0.061M, 0.163M, and 0.368M, respectively.
    After generating \textbf{0.061M} rollout samples, our model achieves the score \textbf{45.47} that exceeds the score \textbf{45.20} of standard GRPO trained for 100 steps with \textbf{0.205M} rollout samples.
    Compared to GRPO (200 Steps), which achieves an average score of \textbf{48.64} with \textbf{0.410M} rollouts, \methodName-GRPO achieves a much higher score of \textbf{50.53} within 200 training steps, i.e., \textbf{0.368M} rollouts, showing a superior final performance with less computational cost.
    
    \item \textbf{(7B) ReMix-PPO v.s., AceReason-Nemotron:} AceReason-Nemotron, the strongest 7B baseline method in our comparison, requires over \textbf{3.584M} rollouts to reach its final score (i.e., \textbf{63.24})\footnote{The score of AceReason-Nemotron is obtained by evaluating the official checkpoint, and the rollout data volume is estimated according to the text and Figure 3 in~\citep{chen2025acereason}.}, whereas \methodName-PPO (50, 75 Steps) achieves a slightly higher accuracy (i.e., \textbf{63.27}, \textbf{64.39}) with \textbf{0.007M}, \textbf{0.011M} rollouts --- over \textbf{a 450x reduction} in rollout data volume.
    
    \item \textbf{(7B) ReMix-PPO v.s., AdaptThink:} AdaptThink, the second strongest 7B baseline method, requires around \textbf{0.307M} rollouts to reach its final score (i.e., \textbf{58.77}), whereas \methodName-PPO (25 Steps) achieves a comparable accuracy (i.e., \textbf{58.49}) with \textbf{0.003M} rollouts --- over \textbf{an 80x reduction} in rollout data volume.
    
    \item \textbf{(7B) ReMix-PPO v.s., PPO:} Compared to PPO (200 Steps) that achieves an average score of \textbf{60.34} with \textbf{0.051M} rollouts, \methodName-PPO achieves a higher score of \textbf{63.27} within 50 training steps, i.e., \textbf{0.007M} rollouts, showing \textbf{a 6x reduction} in rollout data volume.
\end{itemize}

One thing to note is that we found that the average rollout response length of \methodName is lower than the baseline models (the corresponding evidence can be found later in Figure~\ref{fig:res}), hence the exact efficiency should be higher.
This demonstrates the significant superiority of \methodName in improving the efficiency of RFT.
In addition, recall that we use a maximum response generation length of 8,192 tokens in our evaluation (due to computational resource constriant), some of the baseline methods could perform slightly better when a larger length is allowed.

The corresponding detailed factors associated with computational cost for training the 1.5B models and 7B models in the comparison above are shown in Table~\ref{tab:details} and~\ref{tab:details7b} in the appendix, respectively.
Compared to most baselines, our method uses significantly fewer training steps while delivering superior performance. 
Furthermore, our entire training run for \methodName-PPO (1.5B) is executed on a single node with two A800 GPUs over 52 hours, amounting to 104 A800 GPU hours.
This finding shows that SOTA-level gains can be achieved with remarkably reduced compute requirements.

\begin{tcolorbox}[colback=Cornsilk, colframe=lightgray, title=\textbf{Takeaway 1. \methodName can learn a strong reasoning ability in a very efficient manner.}]
\methodName achieves SOTA-level accuracies at 1.5B and 7B scales on five math reasoning benchmarks, with an over 6x to 10x reduction in rollout data volume when outperforming PPO and an over 30x to 450x reduction when performing on par with (or exceeding) the best baseline methods.
\end{tcolorbox}

\subsection{Ablation Studies}
\label{subsec:ablation}

To assess the contribution of each components in our proposed method \methodName, we conduct a series of ablation studies focusing on both training dynamics and final performance.
These experiments are designed to isolate the impact of the three components: Mix-PPG (and increased UTD), Policy Reincarnation, and KL-Convex, within our overall implementation. 
We use \methodName-PPO for the ablation studies in this subsection.

\begin{table}[h]
\centering
\resizebox{\linewidth}{!}{
\begin{tabular}{l|ccccc|c}
\toprule
\textbf{Model} & \textbf{AIME'24} & \textbf{AMC'23} & \textbf{MATH500} & \textbf{Minerva} & \textbf{Olympiad} & \textbf{Avg.}\\\midrule
R1-Distilled-Qwen-1.5B (Base Model) & 33.33 & 43.37 & 67.40 & 16.54 & 27.26 & 37.58 \\
\midrule
PPO (500 Steps) & \underline{36.67} & 62.65 & \textbf{82.60} & 25.73 & 40.14 & 49.56 \\
\rowcolor{Cornsilk} \textbf{\methodName-PPO (350 Steps)} & \underline{36.67} & \textbf{69.88} & 82.00& \textbf{30.15}& 41.78& \textbf{52.10}\\
\midrule
\methodName-PPO w/o UTD & \underline{36.67} & 62.65 & \underline{82.20} & \underline{28.68} & \textbf{42.96} & \underline{50.63} \\
\methodName-PPO w/o KL-Convex & 30.00 & 65.06 & 81.60 & 27.94 & \underline{42.22} & 49.36 \\
\methodName-PPO w/o Policy Reincarnation & 20.00 & \underline{67.47} & 82.00 & 26.84 & 40.00 & 47.26 \\
\methodName-PPO w/o UTD, KL-Convex, Policy Reincarnation & \textbf{40.00} & 57.83 & 80.40 & 25.74 & 39.55 & 48.70 \\
\bottomrule
\end{tabular}}
\vspace{-0.2cm}
\caption{\textbf{Ablation Studies regarding Pass@1 Accuracy.} We focus on the transition from PPO to \methodName-PPO to ablate the components: Mix-PPG, increased UTD, KL-Convex, and policy reincarnation. Note that Mix-PPG is the core of \methodName and the method degenerates to PPO when Mix-PPG is ablated (and other components of \methodName are no longer applicable). 
\textbf{Bolded} and \underline{underlined} values denote the highest and the second-highest scores in each dataset.}
\label{tab:ablation}
\end{table}

The results of the ablation studies regarding Pass@1 accuracy are presented in Table~\ref{tab:ablation}.
First, when Mix-PPG, the core of \methodName, is ablated, the method degenerates to PPO since it does not make sense any longer to apply other components of \methodName.
For the other three components: increased UTD, KL-Convex, and policy reincarnation, ablating each of them leads to a final average score comparable to PPO but lower than \methodName within 500 training steps, even though they all increase the early-stage learning efficiency thanks to Mix-PPG.
Especially when policy reincarnation is not applied, the average score is the worst among the three single-component ablations.
Further, when the three components are removed together, i.e., Mix-PPG works solely, it leads to an even lower score.

This reflects the off-policy nature in Mix-PPG: although it significantly increases the training efficiency, the off-policyness bias may hinder the convergence performance.
This echoes the common sense on the distinction between off-policy RL and on-policy RL in the literature.
The superiority in efficiency brought by Mix-PPG can be observed by referring to the first subplot of Figure~\ref{fig:res}:
Mix-PPG shows a somewhat surprising boost of Pass@1 accuracy within the first 100 training steps, which an increased UTD further enhances it; while the KL-Convex and policy reincarnation in \methodName contribute to the steady asymptotic improvement.

Under the synergy of the proposed techniques, \methodName achieves a quick boost in the early-stage learning, while keeping the asymptotic improvement in later stages.
These findings underscore the effectiveness of our method, which leverages both off-policy and on-policy RL training with an increased UTD ratio during the first stage, followed by the reincarnating on-policy training in the second stage, with KL-Convex constraints applied throughout the entire training process.

\begin{figure}[t]
\vspace{-0.2cm}
    \centering
    \subfigure[Learning Curves for MATH500]{\includegraphics[width=0.975\linewidth]{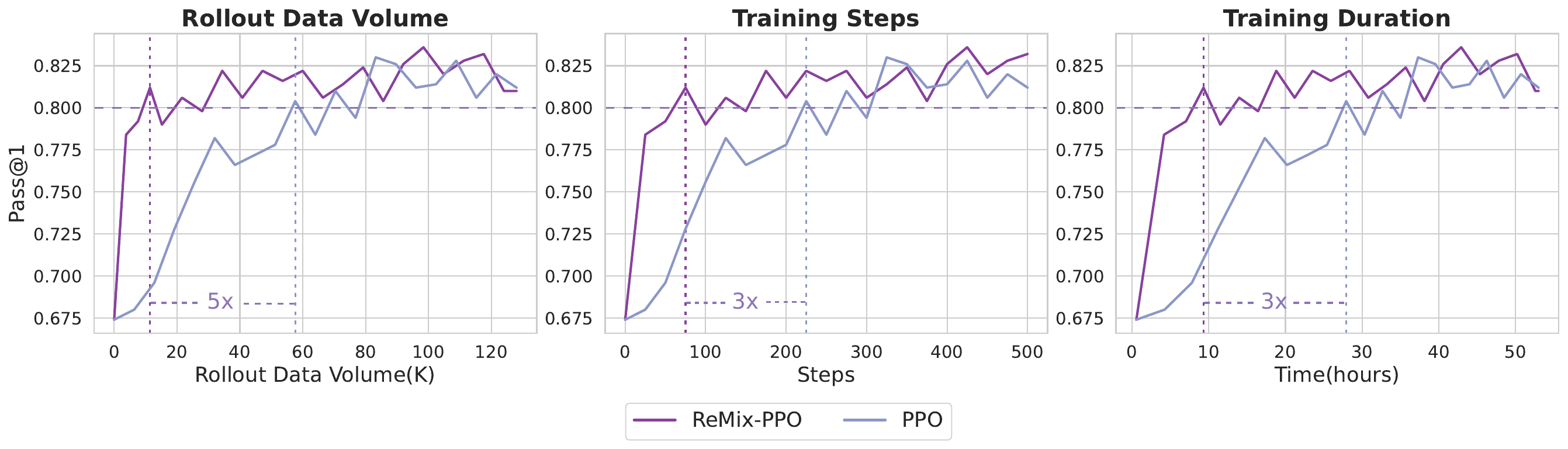}}\\
    \vspace{-0.2cm}
    \subfigure[Learning Curves for Olympiad]{\includegraphics[width=0.975\linewidth]{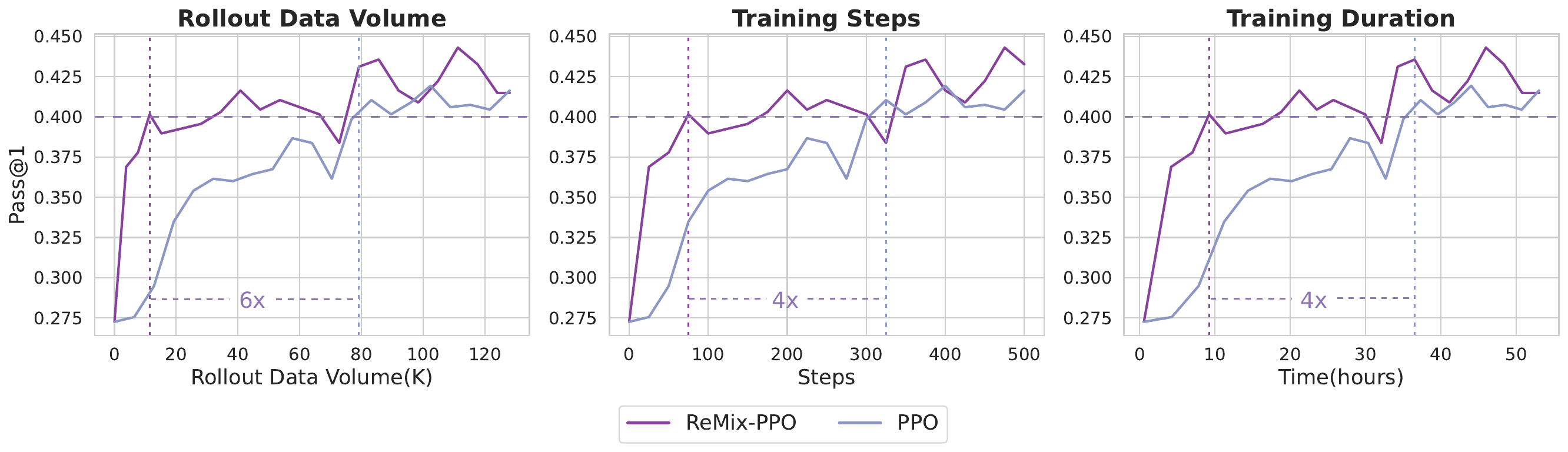}}
    \vspace{-0.3cm}
    \caption{\textbf{Training Efficiency Comparison for \methodName-PPO and PPO (1.5B) on MATH and Olympiad.} We evaluate training efficiency across three dimensions: rollout data volume, training steps, and training duration. \methodName achieves a score above 40\%, around \textbf{3x to 6x faster} than PPO.}
    \label{fig:ppovsours}
\vspace{-0.3cm}
\end{figure}

\vspace{-0.5cm}
\paragraph{Training Curves}
In addition to the efficiency evaluation in terms of rollout data volume, we present the training curves for \methodName-PPO and PPO in Figure~\ref{fig:ppovsours} on MATH500 and Olympiad regarding two more efficiency aspects, i.e., training steps and wall-clock time.
Note that training steps do not reflect the training cost fairly since different models use various batch sizes and rollout settings (see the details in Table~\ref{tab:details} and~\ref{tab:details7b}).
Our method demonstrates superior training efficiency by achieving a score above 80 on MATH500 with a 3x and 5x reduction in training steps and wall-clock time, and achieving a score above 40 on Olympiad with a 4x reduction similarly.
Notably, we can observe that \methodName exhibits a prominent boost in the early stage of training, which emphasizes the significance of Mix-PPG and the increased UTD ratio for improving training efficiency again.
We provide more training curves for the other three benchmarks in Appendix~\ref{app:training_curves}.

\begin{tcolorbox}[colback=Cornsilk, colframe=lightgray, title=\textbf{Takeaway 2. The components in \methodName work in synergy for both efficiency and final performance.}]
Mix-PPG with an increased UTD boosts early-stage training significantly, while policy reincarnation plays a critical role to ensure asymptotic improvement.
\end{tcolorbox}

\subsection{Various Analysis}
\label{subsec:analysis}

In this subsection, we present a multifaceted empirical analysis to gain better understanding of the effects of our proposed bb method. For convenience, we use \methodName-PPO for the analysis in the following.

\subsubsection{The Impact of Historical Sample Reuse}
\label{subsubsec:onoffratio}

\begin{figure}[h]
\vspace{-0.2cm}
    \centering
    \includegraphics[width=\linewidth]{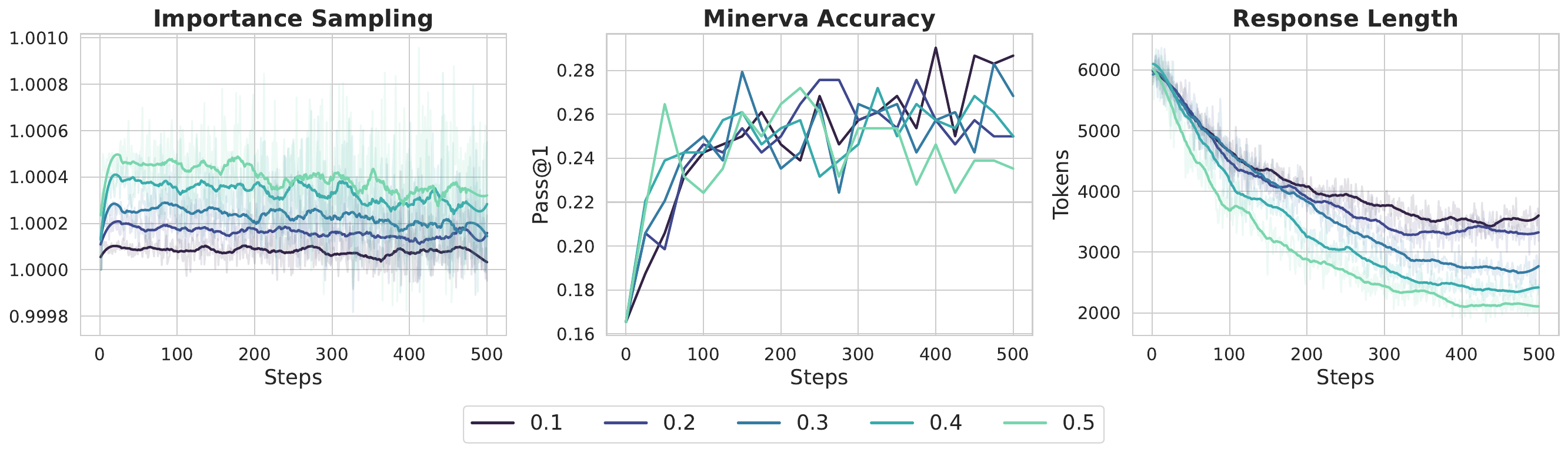}
    \vspace{-0.6cm}
    \caption{\textbf{Training Dynamics regarding Importance Sampling Ratio, Accuracy, and Response Length under Varying Proportions of Off-policy Data $p$ for Mix-PPG.} Leveraging more off-policy data leads to a larger policy distribution shift, a faster early boost in accuracy yet worse later-stage performance, and a shorter response length.}
    \label{fig:onoffratio}
\end{figure}

Lying in the center of our method, Mix-PPG leverages both the historical off-policy data generated during the past training process and the conventional on-policy data.
To investigate the impact of off-policy data in Mix-PPG, we conduct an empirical analysis by varying the proportion of off-policy data $p \in \{0.1, 0.2, 0.3, 0.4, 0.5\}$ (the UTD ratio is set to 1 here for isolation).
In addition to the Pass@1 accuracy, we use two more metrics: the importance ratio $\frac{\pi_k}{\pi_{k-i}}$ that quantifies the distributional shift between current and historical policies, and the response length that reflects the reasoning behavior of the model.
The results are shown in Figure~\ref{fig:onoffratio}.

We can observe that as the increase of off-policy data proportion, the distribution shift between current and historical policies measured by 
the importance sampling ratio also increases. This echoes the common knowledge in the RL community on the impact of off-policyness on policy training. 
Meanwhile, by referring to the training curves of Pass@1 accuracy, we can observe that larger off-policy data proportions (e.g., 0.4 and 0.5) facilitate faster initial gains, suggesting improved sample efficiency in early training stages. 
However, these configurations exhibit inferior final performance or even degradation in the later stages of training. 
This observation aligns with prior insights, showing that too much off-policyness can destablize the training process and cripple the convergence.
This indicates that a balanced off-/on- ratio is essential to benefit from the superior training efficiency and avoid instability due to excessive policy divergence.

\begin{tcolorbox}[colback=Cornsilk, colframe=lightgray, title=\textbf{Takeaway 3. A trade-off between efficiency and final performance needs to be balanced when incorporating off-policyness.}]
For \methodName, leveraging more historical off-policy data leads to a more noticeable boost in early-stage training, while exacerbating the instability and being more prone to degradation in later stages.
\end{tcolorbox}

Moreover, we found that higher off-policy data proportions consistently lead to lower response lengths. This reduction in length may contribute to the observed drop in final reasoning performance for these settings. We further analyze this effect in Section~\ref{subsubsec:reslengthonoff} below.

\subsubsection{How Off-Policy Learning Affects Reasoning Behaviors}
\label{subsubsec:reslengthonoff}

Beyond the evidence observed in Figure~\ref{fig:onoffratio}, we further the study on the relationship between off-policy RL imposed by \methodName and the reasoning behaviors during the learning process of LLMs.

\begin{figure}[h]
    \centering
    \includegraphics[width=\linewidth]{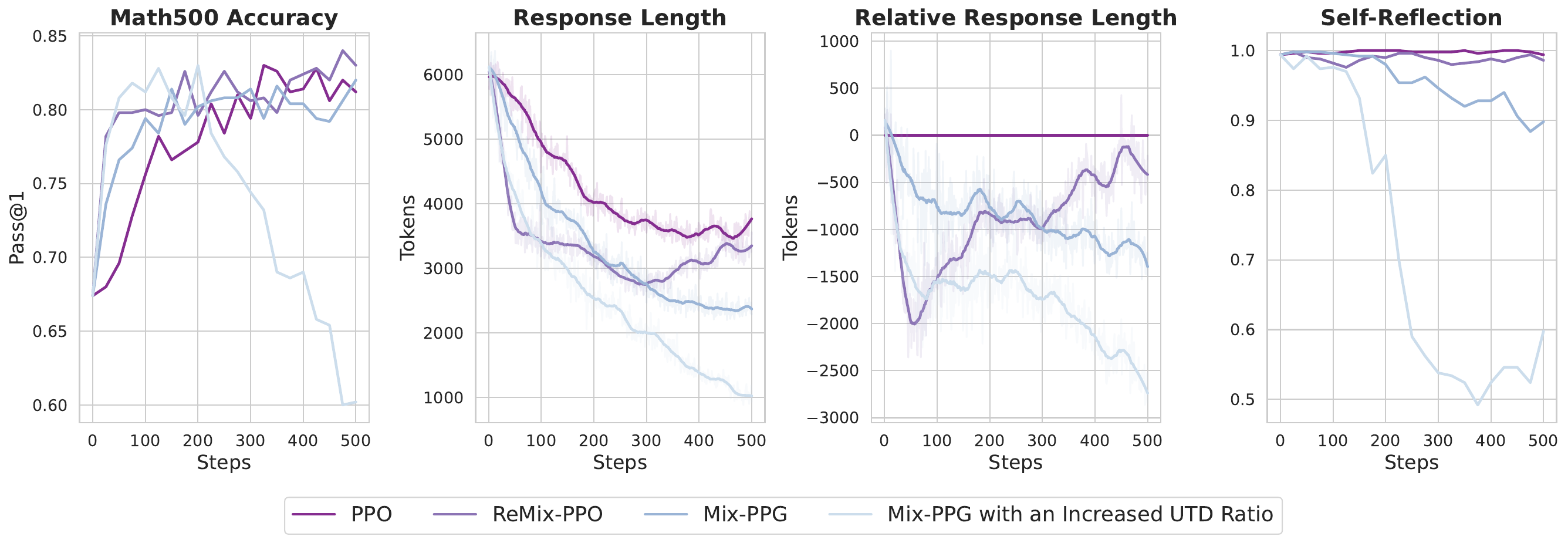}
    \vspace{-0.7cm}
    \caption{\textbf{Training Dynamics regarding Accuracy, Response Length, Self-reflection Rate for On-policy v.s. Off-policy Training.} 
    Mix-PPG significantly increases the early-stage training efficiency, while showing a rapid decrease in response length and self-reflection rate.
    An increased UTD ratio further enhances the efficiency, but results in a severe degradation in accuracy.
    \textbf{\methodName shows a merged learning behavior and perfectly combines the superior efficiency and the asymptotic improvement}, thanks to the policy reincarnation.}
    \label{fig:res}
\end{figure}

In addition to accuracy and response length, we make use of two more metrics:
relative response length (against the training dynamics of PPO), and self-reflection rate that is calculated according to the occurrence of reflection tokens (e.g., `verify', `re-examine', `check', `but', `wait', `confirm', etc.). 
Moreover, we compare PPO, \methodName-PPO (i.e., \textit{Ours}), Mix-PPG and Mix-PPG with an increased UTD (as previously presented in Table~\ref{tab:ablation}).
The results are shown in Figure~\ref {fig:res}.

The vanilla PPO shows a steady increase of Pass@1 accuracy as well as a decrease in response length.
PPO consistently elicits self-reflection behaviors from the base model and keeps the self-reflection rate near 1.
Mix-PPG significantly increases the training efficiency before 100 training steps, showing a more rapid decrease in response length and a notable decrease in self-reflection rate.
Meanwhile, it also leads to inferior peak performance in the later stages of training.
When applying an increased UTD ratio, Mix-PPG exhibits an even higher efficiency but shows a quick degradation after 200 training steps.
This is accompanied by the quick decrease in response length and self-reflection rate.
A rapid reduction in response length leads to a diminished capacity for reflective behavior in the model. 
This trend suggests that the model no longer engages in a step-by-step reasoning process, consequently reducing opportunities for trial-and-error. Such a shift can prematurely drive the model to generate a final answer without adequate intermediate deliberation, resulting in a drop of accuracy. Please see Appendix~\ref{app:case} for detailed cases. 

More interestingly, \methodName seems to perfectly combine the superior efficiency of Mix-PPG with an increased UTD in the early stage and the asymptotic improvement of PPO, thanks to the existence of policy reincarnation.
From the angles of response length and self-reflection rate, \methodName also exhibits \textbf{a merged dynamic pattern of reasoning behaviors}: \methodName first quickly decreases the response length and suppresses frequent self-reflection to improve its accuracy in the early stage of training; it then expands the responses and uses more reflection for careful exploration and further improvement of accuracy.

\begin{tcolorbox}[colback=Cornsilk, colframe=lightgray, title=\textbf{Takeaway 4. Off-policy learning leads to a suppression of self-reflection behaviors and thus cripples reasoning ability.}]
Incorporating more off-policyness leads to a quicker decrease in response length and a more severe drop in self-reflection rate, leading to severe degradation of accuracy.
\methodName benefits from the superior efficiency of off-policy learning while suffers no degradation.
\end{tcolorbox}

\vspace{-0.3cm}
\paragraph{The Implicit Preference of Off-policy Learning for Shorter Responses} To gain a more thorough insight into why off-policy learning leads to the observed reasoning behaviors, we conduct a formal analysis on the learning dynamics when optimizing the Mix-PPG loss function $L_k^{\textcolor{blue}{\text{Mix-PPG}}}(\theta)$ (shown in Eq.~\ref{eq:mix_ppg}).
The empirical evidence in Figure~\ref{fig:onoffratio} shows that the importance sampling ratio is close to 1, thus we simplify our formal analysis by ignoring the clip mechanism in $L_k^{\textcolor{blue}{\text{Mix-PPG}}}(\theta)$.
Similar to the transformation presented in~\citep{Fatemi25Concise}, the average loss of Mix-PPG can be formulated below:
\begin{equation}
    L_{\text{Avg}}^{\textcolor{blue}{\text{Mix-PPG}}} = \frac{1}{H}\sum_{h=0}^{H} L_{h}^{\textcolor{blue}{\text{Mix-PPG}}} \propto - \frac{1}{H}\sum_{h=0}^{H} \textcolor{blue}{r_{\theta}^{k-1}} A_h^{\pi_k}
\label{eq:off_analysis}
\end{equation}
Based on the equation above, we can find: when the advantage estimate is \textit{negative}, the model learns to minimize the loss by steering its policy to achieve a lower importance sampling ratio.

First, let us validate the sign of actual advantage estimates. By referring to the policy loss curve shown in Figure~\ref{fig:policy_loss}, we can observe that the policy loss is almost always positive, which means the advantage estimates are negative most of the time. This matches the case we mentioned above.
Therefore, the learning dynamics of reasoning behaviors should be explained by the policy optimization behavior towards a lower importance sampling ratio for loss minimization.
Since the average loss is computed based on the data distribution of historical policy $\pi_{k-i}$, there apparently exists a \textit{Whipping Effect}: the longer the response is, the larger the distribution shift should be on later states.
Consequently, the model tends to prefer shorter responses in order to reduce the average loss associated with long rollout trajectories.
This tendency will be further amplified as the proportion of off-policy data increases.

As shown in Figure~\ref{fig:res}, leveraging more off-policy data leads to a larger distribution shift measured by the importance sampling ratio, inducing a larger loss when the advantage is negative.
This further results in a larger gradient that steers the policy to shorten the response length.
This shortening of response length suppresses the reflective reasoning behaviors, thus inducing a degradation of final performance.
A consistent slight decrease of the importance sampling in Figure~\ref{fig:res} can also be explained by the shortening of response length, as the whipping effect gradually diminishes.

\subsubsection{The Performance under Constrained Maximum Response Length}

Since \methodName shows a feature in generating more concise responses as discovered above, we conduct an additional experiment to evaluate the performance of our model when the maximum response length is constrained.
Different from the default evaluation setting of 8,192 maximum response length, we halve the maximum response length to 4,196 tokens for \methodName during evaluation.
For comparison, we evaluate \methodName-PPO (1.5B) with the base model, DeepScaleR, and PPO under the halved maximum response length.
The results are summarized in Table~\ref{tab:responselength}.

\begin{table}[t!htbp]
\centering
\resizebox{\linewidth}{!}{
\begin{tabular}{l|ccccc|c}
\toprule
\textbf{Model} & \textbf{AIME'24}  & \textbf{AMC'23} & \textbf{MATH500} & \textbf{Minerva} & \textbf{Olympiad} & \textbf{Avg.} \\
\midrule
\multicolumn{7}{c}{Maximum response length: \textcolor{red}{4096} tokens}\\\midrule
R1-Distilled-Qwen-1.5B (Base Model) & \underline{20.00} & 37.35 & 60.40 & 13.24 & 22.37 & 30.67\textsubscript{\textcolor[rgb]{0,0.4,0}{↓6.91}} \\
DeepScaleR & 10.00 & \underline{49.4} & 75.00 & 21.32 & 34.22 & 37.99\textsubscript{\textcolor[rgb]{0,0.4,0}{↓14.15}} \\
PPO (500 Steps) & \underline{20.00} & 48.19 & \underline{77.60} & \underline{25.00} & \underline{38.96} & \underline{41.95}\textsubscript{\textcolor[rgb]{0,0.4,0}{↓7.61}}  \\
\midrule
\rowcolor{Cornsilk}
\textbf{\methodName-PPO (350 Steps)} & \textbf{23.33} & \textbf{59.04} & \textbf{79.00} & \textbf{27.57} & \textbf{39.11} & \textbf{45.61}\textsubscript{\textcolor[rgb]{0,0.4,0}{↓6.49}}\\
\bottomrule
\end{tabular}}
\vspace{-0.2cm}
\caption{
\textbf{Performance Evaluation of 1.5B Models with 4k Maximum Response Length.} 
The arrow \textcolor[rgb]{0,0.4,0}{↓} denotes the accuracy degradation compared to the results with 8k maximum response length (referring to the results in Table~\ref{tab:1.5b}).
All the models are negatively influenced by the halved maximum response length.
Compared with DeepScaleR, \textbf{\methodName-PPO exhibits the smallest decrease in model performance and performs the best}, thanks to its concise and shorter reasoning behaviors.}
\label{tab:responselength}
\end{table}

All the models are negatively influenced by the halved maximum response length, which matches the intuition.
Notably, DeepScaleR, the best 1.5B baseline model used in our work, suffers a significant performance drop when the maximum response length is limited to 4,192 tokens.
In contrast, \methodName-PPO exhibits the smallest decrease in model performance and performs the best in this constrained setting.
This finding underscores the resilience of the preference for a concise and shorter reasoning process learned via off-policy training of \methodName in handling the constraints on response length.

\begin{tcolorbox}[colback=Cornsilk, colframe=lightgray, title=\textbf{Takeaway 5. \methodName favors concise reasoning and is resilient to constraints on response length.}]
The constraint on maximum response length greatly degrades the test-time reasoning performance of LLM models, while \methodName suffers less thanks to its preference for a concise reasoning process.
\end{tcolorbox}

\subsubsection{The Impact of Guide Tokens in Prompt Template}

In addition, we investigate the critical role of the prompt template for response generation used during training and evaluation.
To establish a comparison, we make use of a prompt template \textit{without guide tokens} (as shown below).
Recall the standard prompt template we presented in Section~\ref{subsec:setups}, the difference is that the prompt template without guide tokens does not contain the guide tokens that appear as the prefix (i.e., \texttt{<begin\_of\_sentence><|User|>}) and the suffix (i.e., \texttt{<|Assistant|><think>}).
\begin{tcolorbox}[colback=white, colframe=black, title=System Prompt (Without Guide Tokens)]
\textcolor{blue}{\sout{\textless\textbar begin\_of\_sentence\textbar\textgreater\textless\textbar User\textbar\textgreater}}Xenia and Sergey play the following game. Xenia thinks of a positive integer $N$ not exceeding 5000. Then she fixes 20 distinct positive integers $a_{1}, a_{2}, \ldots, a_{20}$ such that, 
for each $k=1,2, \ldots, 20$, the numbers $N$ and $a_{k}$ are congruent modulo $k$. By a move, Sergey tells Xenia a set $S$ of positive integers not exceeding 20, 
and she tells him back the set $\left\{a_{k}: k \in S\right\}$ without spelling out which number corresponds to which index. How many moves does Sergey need to determine for sure the number Xenia thought of? \textcolor{blue}{Let's think step by step and output the final answer within \text{\textbackslash boxed\{\}}.\ \sout{\textless\textbar Assistant\textbar\textgreater\textless think\textgreater}}
\end{tcolorbox}

To investigate the impact of different prompt templates, we define a response as format-correct if it includes content enclosed within paired \texttt{<think>} \texttt{</think>}.
Parallel to the 1.5B base model and \methodName-PPO, we consider a variant of \methodName-PPO that is trained without guide tokens, denoted as \methodName-PPO w/o Guide Tokens.
We evaluate the performance of the candidate models in terms of Pass@1 accuracy\footnote{Note that the correct answer with a wrong format is still counted as correct for Pass@1 accuracy here.} and format correctness on MATH500, when using the standard template (i.e., with guide tokens) and the modified template without guide tokens.
The purpose of this experiment is to answer two questions: (1) whether the models trained with guide tokens (i.e., the base model, \methodName-PPO) can also perform well when the guide tokens are not prompted during evaluation; (2) whether the model trained without guide tokens can also obey the format and output the solution.
The results are summarized in Table~\ref{template}.

\vspace{-0.1cm}
\begin{table}[h]
\centering
\resizebox{\linewidth}{!}{
\begin{tabular}{l|cc|cc|cc}
\toprule
\textbf{Model} & \multicolumn{2}{c}{\textbf{Eval w/ Standard Temp.} (\textcolor{red}{↑})} & \multicolumn{2}{c}{\textbf{Eval w/o Guide Tokens} (\textcolor{red}{↑})} & \multicolumn{2}{c}{\textbf{Relative Decrease} (\textcolor[rgb]{0,0.4,0}{↓})} \\
 & \textbf{Pass@1}  & \textbf{Format Cor.}  & \textbf{Pass@1}  & \textbf{Format Cor.} & \textbf{Pass@1}  & \textbf{Format Cor.}\\\midrule
R1-Distilled-Qwen-1.5B (Base Model) & 67.40 & 70.00 & 52.00 & 0 & 15.40 & 70.00 \\
\rowcolor{Cornsilk}
ReMix-PPO~(350 Steps) & 82.00 & 93.60 & 71.00 & 0 & 11.00 & 93.60 \\
\rowcolor{Cornsilk}
ReMix-PPO w/o Guide Tokens (500 Steps)& 82.00 & 92.20 & 77.60 & 91.20 & 4.40 & 1.00 \\
\bottomrule
\end{tabular}}
\vspace{-0.2cm}
\caption{\textbf{Performance Evaluation of 1.5B Models with and without Guide Tokens on MATH500.} 
Both the base model and \methodName-PPO show a 0 format correctness when the guide tokens are missing during evaluation, while \methodName-PPO exhibits a smaller drop in the accuracy.
For the variant of \methodName trained without the guide tokens, it \textbf{performs well under both the two template settings}.
\textcolor{red}{↑} means higher is better and \textcolor[rgb]{0,0.4,0}{↓} means lower is better.
}
\label{template}
\end{table}
\vspace{-0.2cm}

The results show that the base model yields a format correctness of 0 when evaluated without the guide tokens,
accompanied by a decrease of 15.40 points in Pass@1 accuracy.
Similarly, \methodName-PPO also exhibits a 0 format correctness yet a smaller decrease of 11.00 in the accuracy.
This indicates that the presence of the \texttt{<think>} token in the prompt helps the model to autonomously generate a closing \texttt{</think>} tag, maintaining format consistency.
Thus, it delivers a negative answer to the first question above, while \methodName shows a better robustness to the absence of the guide tokens.

In contrast, the variant trained without the guide tokens also performs well when using the standard template, and achieves an increase from 77.60 to 82.00, reaching the same performance as \methodName that is trained with the guide tokens explicitly.
It also maintains consistently high format correctness.
This shows a good robustness to prompt change.
We found similar results for the other four math reasoning tasks as well.

We hypothesize that removing the guide tokens during the training of \methodName allows the model to explore a broader distribution, rather than overfitting to the explicit guide tokens in the standard template. 
Such flexibility encourages the model to internalize reasoning behavior in a robust and general manner, instead of relying on external structural cues too much.
As a result, it becomes more robust to prompt variation at inference time. The smaller relative degradation observed in both accuracy and format correctness supports this view.

\begin{tcolorbox}[colback=Cornsilk, colframe=lightgray, title=\textbf{Takeaway 6. \methodName is more robust to the variation of prompt template.}]
Removing explicit guide tokens in the standard template significantly cripples the performance of the base model, while \methodName exhibits better robustness and compatability to the absence of the guide tokens during both training and evaluation.
\end{tcolorbox}

\section{Related Work}
\label{sec:related_work}
Post-training enhancement of LLM reasoning capabilities predominantly follows two paradigms~\citep{li2025system}. The first, inference-time optimization, improves reasoning without updating model parameters through techniques like Chain-of-Thought (CoT) prompting~\citep{wei2022chain}, parellel reasoning and itegration~\citep{wang2022self}, self-reflection~\citep{ji2023towards}, tree-based search~\citep{zhang2024rest}, and macro-action-guided cognitive reasoning~\citep{liu2025chaos}. Despite their effectiveness, the performance of these methods is fundamentally constrained by the model's inherent capabilities.
The second paradigm, parameter fine-tuning, aims to enhance these intrinsic abilities into LLM. While SFT on high-quality reasoning data is a common approach, its effectiveness is often limited by data availability and scalability~\citep{zelikman2024star}. Consequently, RLVR has emerged as a powerful alternative, learning directly from reward signals to unlock superior performance, as demonstrated by models like DeepSeek-R1~\citep{guo2025deepseek}. Notably, this differs from preference-based RL which learns from a reward model trained on human/AI feedback~\citep{bai2022training,bai2022constitutional,liu2024enhancing}, as the RLVR here utilizes direct, verifiable reward signals.
Our work is situated within the RFT paradigm, especially under varifirable reward.

The majority of existing RFT research has relied on on-policy RL algorithms prized for their training stability, such as PPO~\citep{SchulmanWDRK17PPO}. Some recent approaches have sought to improve efficiency by modifying the RL architecture (e.g., GRPO~\citep{Shao24GRPO}) or relaxing optimization constraints~\citep{seed2025seed1}. However, these on-policy RL methods exhibit severe sample inefficiency, as they require fresh samples for each iteration of gradient updates. 
To alleviate this, recent research has begun to incorporate off-policy data in RL training. 
\cite{tang2025rl} propose AGRO for a unified algorithm to leverage any-generation data, encompassing both on- and off-policy samples. However, their experimental results show that off-policy training is inferior to on-policy training, underscoring the non-trivial challenge of achieving stable and effective off-policy training for LLMs.
Tapered Off-Policy REINFORCE~\citep{roux2025tapered} introduces a novel variant of importance sampling to downweight negative trajectories that are not likely under the current policy, while allowing positive trajectories to be upweighted.
This enables the utilization of both off-policy and on-policy rollout trajectories.
The method is trained and evaluated on GSM8K and MATH, leaving its efficacy on broader reasoning tasks unknown.

Recently, concurrent to our work, 
Based on REINFORCE, AsymRE~\citep{arnal2025asymmetric} is proposed to leverage both off-policy and on-policy data by introducing a tunable baseline. An asymmetry is presented that while on-policy updates safely leverage both positive and negative signals, off-policy updates benefit more from positive rewards, which to some extent echoes the idea proposed in~\citep{roux2025tapered}. AsymRE is trained and evaluated on MATH.
RePO~\citep{li2025repo} is proposed upon GRPO to replay both historical off-policy data and on-policy data together during typical GRPO training.
Different off-policy data replay strategies are studied, among which recency-based and reward-based strategies show improved performance. The RePO models are trained with a maximum response length of 1,024, thus showing limited performance on math reasoning benchmarks.
By following the principle of Soft RL, SPO~\citep{cohen2025soft}) is proposed to leverage both off-policy and on-policy data based on Cumulative Q-Parameterization.
SPO is trained and evaluated for code contests and demonstrates superior performance to the standard PPO.
In contrast, LUFFY~\citep{yan2025learning} uses off-policy samples from superior models (e.g., DeepSeek-R1) and employing policy shaping. However, in essence, this is more akin to learning from demonstrations rather than the canonical off-policy RL where the behavior policy is often one of the historical policies or a separate inferior policy. Moreover, the idea of off-policy guidance is orthogonal to our method.

While early efforts have conducted first-step explorations on realizing off-policy learning for RFT, they have primarily focused on adapting existing on-policy methods (e.g., PPO, GRPO, REINFORCE) to off-policy data from the angles of modifying importance sampling, leveraging data or trajectories asymmetrically, etc. 
These initial steps have not investigated the essential effects of off-policy learning on reasoning behaviors, while leaving the potential of existing off-policy RL techniques unexplored. 
In the broader field of RL, methods such as Rainbow~\citep{HesselMHSODHPAS18RAINBOW}, TD3~\citep{Fujimoto2018TD3}, and SAC~\citep{haarnoja2018soft} have set a precedent for leveraging historical data to improve sample efficiency. Building on this, advanced research has pursued maximizing data utilization through high UTD ratios, managing the resultant estimation errors with techniques like ensemble learning, as seen in REDQ~\citep{chenrandomized}, DroQ~\citep{hiraoka2021dropout}, and SPEQ~\citep{romeospeq}. Concurrently, novel approaches have emerged, including hybrid methods that seek an optimal balance between the stability of on-policy learning and the efficiency of off-policy methods~\citep{QueeneyPC21Geppo}, as well as fully offline algorithms designed to mitigate extrapolation errors from static datasets~\citep{ma2024iteratively,ma2023reining}. 
The value of ReMix lies in its departure from simply implementing off-policy RL in the context of RFT. Instead, by drawing inspiration from rich RL literature, our research aims to conduct an in-depth investigation of different off-policy RL techniques and integrate them to improve the RFT process effectively, thereby significantly enhancing the efficiency and performance of LLM fine-tuning.

\section{Conclusion}
\label{sec:conclusion}

In this paper, we aim to address the notorious drawback of on-policy RFT methods (e.g., PPO and GRPO) on training inefficiency and prohibitive computational cost.
We launch the renaissance of
off-policy RL and propose Reincarnating Mix-policy Proximal Policy Gradient (\methodName), a general approach
to enable on-policy RFT methods like PPO and GRPO to leverage off-policy data.
In our experiments, we implement \methodName upon PPO, GRPO, and 1.5B-, 7B-scale base models.
Through evaluating the reasoning accuracy and training efficiency of \methodName on five math reasoning benchmarks against 15 recent advanced baseline models, we demonstrate the superiority of \methodName in improving training efficiency and achieving SOTA-level reasoning performance with a great reduction in training cost.

\vspace{-0.5cm}
\paragraph{Limitations} Due to the computational resource constraint, we did not conduct experiments on models larger than 7B, which leaves the practical scaling performance of our proposed method untested thoroughly. 
For the utilization of off-policy data, we use fixed proportions in this work, although we believe an adaptive control on the proportion of off-policy data should be possible and favorable.
Moreover, our method is orthogonal to many of the advanced RFT methods considered and not considered in our experiments, while we do not explore the combination of them. We believe that integrating off-policy learning and other advanced techniques is promising to realize new LLM models that are more efficient and powerful at the same time. 
We leave these potential angles for the future.

\clearpage
\bibliography{main}

\appendix
\newpage

\appendix

\section{Advantage Estimation}

To enable stable off-policy training, we adopt a V-trace~\citep{EspeholtSMSMWDF18Impala} formulation for generalized advantage estimation (GAE)~\citep{SchulmanMLJA15GAE}, which incorporates truncated importance sampling ratios to correct for policy mismatch. We first compute the temporal-difference error(TD-error) at each time step \( t \) as
\begin{align}
\delta^V_t = r(s_t, a_t) + \gamma V(s_{t+1}) - V(s_t),
\end{align}
and define the truncated importance sampling weight \( c_t = \min\left(\bar{c}, \frac{\pi_k(a \mid s)}{\pi_{k-i}(a \mid s)}\right) \), where \( \bar{c} \) is a clipping threshold to limit the variance of the correction, we use \( \bar{c} \) = 1 in our implement.

The advantage at step \( t \) is estimated recursively using the V-trace correction as
\begin{align}
A_t = \delta^V_t + \gamma \lambda c_t A_{t+1},
\end{align}
and the return-to-go is computed by combining the advantage estimate with the baseline value:
\begin{align}
\text{RTG}_t = A_t \cdot c_t + V(s_t).
\end{align}

This V-trace corrected GAE formulation ensures that the estimated advantages remain stable and consistent under significant off-policy drift, which is critical in our training regime involving long-horizon trajectories and evolving policies.

\section{A Brief Overview of Baseline Models}
\label{app:baseline_intro}
\subsection{1.5B Models}
\begin{itemize}
    \item \textbf{Open-RS Series}~\citep{dang2025reinforcementopenrs}:
    The Open-RS series employs the GRPO algorithm to train language models, using datasets constructed by filtering and combining existing corpora. Specifically, \textbf{Open-RS1} utilizes dataset with 18,615 samples with accuracy and format rewards, \textbf{Open-RS2} incorporates dataset with 7,000 samples and shorter maximum response length while retaining the same reward functions. Compared to \textbf{Open-RS2}, \textbf{Open-RS3} replaces the accuracy reward with a cosine reward and adds an English-only instruction to the system prompt.
    \item \textbf{DeepScaleR}~\citep{deepscaler2025}:
    DeepScaleR is obtained via a two-phase training process with the GRPO algorithm: starting with 8k context for efficient reasoning, then scaling up to 16k and 24k contexts to address more challenging problems.
    \item \textbf{II-Thought}~\citep{II-Thought}:
    Based on a systematic analysis of existing public datasets, the authors constructed a large-scale, high-quality dataset comprising over 300,000 reasoning problems across multiple domains. Each sample was rigorously filtered and deduplicated. Subsequently, the models were trained on this curated dataset, using the GRPO algorithm.
    \item \textbf{FastCuRL Series}~\citep{song2025fastcurl}:
    The FastCuRL Series adopts a multi-stage training process where both context length and data complexity (defined by input prompt length) are progressively increased. Training starts with short-context and low-complexity data, then moves to longer contexts with medium and high-complexity datasets.
    \item \textbf{L1 Series}~\citep{aggarwal2025l1}:
    The L1 Series trains models using Length-Controlled Policy Optimization (LCPO), a method that encourages correct answers while matching a target output length specified in the prompt (measured by input prompt length). \textbf{L1-Exact} enforces exact-length generation by penalizing deviation from the target length, while \textbf{L1-Max} applies a soft maximum-length constraint, allowing shorter outputs when appropriate but discouraging overruns.
    \item \textbf{AdaptThink}~\citep{zhang2025adaptthink}:
    AdaptThink is an RFT method that trains reasoning models to choose between two modes --- Thinking and NoThinking --- based on problem difficulty. It uses a constrained optimization objective to encourage NoThinking while maintaining performance, and an importance sampling strategy to balance both modes during training.
\end{itemize}

\subsection{7B Models}
\label{app:7b_model_intro}
\begin{itemize}
    \item \textbf{ReasonFlux-F1}~\citep{yang2025reasonflux}:
    ReasonFlux-F1 is an SFT model obtained by finetuning an R1-Distill model\footnote{\url{https://github.com/Gen-Verse/ReasonFlux/blob/main/ReasonFlux_F1/README.md}} based on template-augmented reasoning trajectories collected by ReasonFlux-v1. These trajectories are first enhanced with structured templates, then transformed into a long chain-of-thought format.
    \item \textbf{Light-R1}~\citep{lightr1proj}:
    Light-R1 is a multi-stage post-training framework. It begins with curriculum-based supervised fine-tuning (SFT) using progressively harder data, followed by Direct Preference Optimization (DPO) and an RFT process with GRPO on a filtered dataset. \texttt{Light-R1-7B-DS} is trained only in the second SFT stage of the framework. Thus, the Light-R1 7B baseline model used in our experiments is an SFT model rather than an RFT model.
    \item \textbf{Skywork-OR1-Preview}~\citep{he2025skywork}:
    Skywork-OR1-Preview is trained on a curated dataset of math and coding problems, selected through model-aware difficulty estimation and quality filtering. The training process modifies GRPO by incorporating both offline and online difficulty-based filtering, rejection sampling, and a multi-stage curriculum with adaptive entropy control.
    \item \textbf{Polaris}~\citep{Polaris2025}:
    Polaris adopts a multi-stage RL training approach with careful data difficulty control, using a data distribution with a slight bias toward challenging problems and dynamically adjusting question difficulty during training. It initializes sampling temperature based on rollout diversity and gradually increases it during training. It employs length extrapolation techniques, enabling longer CoT generation at inference while keeping training rollouts short.
    \item \textbf{AdaptThink}~\citep{zhang2025adaptthink}:
    The methodology for the AdaptThink 7B model is identical to that of the AdaptThink 1.5B model, as previously described.
    \item \textbf{AceReason-Nemotron}~\citep{chen2025acereason}:
    AceReason-Nemotron adopts the GRPO algorithm without KL divergence and avoids entropy collapse through controlled updates. The model is first trained on math-only prompts, then on code-only prompts, following a curriculum with progressively increasing response lengths.
\end{itemize}

\section{Training Details}
\label{app:trainingde}

\paragraph{Hyperparmeters} The major hyperparameter choices are shown in Table~\ref{tab:hypers}.

\begin{table}[h]
\centering
\begin{tabular}{l|l}
\toprule
\textbf{Parameter} & \textbf{Value} \\
\midrule
\multicolumn{2}{l}{\textit{Training Configuration}} \\
\midrule
critic\_warmup & 0 \\
learning\_rate & 1e-6 \\
clip\_ratio & 0.2 \\
lam & 1\\
tau & 0.95\\
entropy\_coeff & 0.001\\
clipping\_gradient & true \\
do\_sample & true \\
test\_freq & 25 \\
\midrule
\multicolumn{2}{l}{\textit{Training Configuration for ReMix-GRPO and GRPO}} \\
\midrule
kl\_loss\_coef & 0.001 \\
kl\_loss\_type & low\_var\_kl \\
n (gen per prompt) & 8\\
\bottomrule
\end{tabular}
\caption{\textbf{Hyperparameter setups for PPO, GRPO and ReMix trainer.}}
\label{tab:hypers}
\end{table}

\paragraph{Comparison of Training Detail on Computational Cost for 1.5B Models}
The corresponding detailed factors associated with computational cost for training the 1.5B models in the comparison above are shown in Table~\ref{tab:details}.
Compared to most baselines, our method uses nearly half the number of training steps (500 v.s. $\geq$ 860) while delivering superior performance. 
Furthermore, our entire training run is executed on a single node with just two A800 GPUs over 52 hours, amounting to 104 A800 GPU hours.
This finding shows that state-of-the-art gains can be achieved with markedly reduced compute requirements.

\begin{table}[t!htbp]
\centering
\resizebox{\linewidth}{!}{
\begin{tabular}{l|cccccc}
\toprule
\textbf{Model} & \textbf{Traing Steps} & \textbf{Rollout Batch Size} & \textbf{Gen per Prompt} & \textbf{Max Responses Length} & \textbf{Number of GPUs} \\\midrule
DeepScaleR & 1750 steps & \textit{128,128,128} & \textit{8,16,16} & 8k,16k,24k & 8,32,32 \\
FASTCuRL-preview & 860 steps & 128,64,64,64 & 8,8,8,16 & 8k,16k,24k,16k & 8 \\
FASTCuRL-v3 & 2620 steps & \textit{128,64,64,64,64} & \textit{8,8,8,16,16} & 8k,16k,24k,16k,16k & 8 \\
II-Thought & - & \textit{1024} & \textit{5} & 32k & \textit{8} \\
adapt\ think & 314 steps & 128 & 16 & 16k & 8 \\
Open-RS1 & 100 steps & 96 & 6 & 4k & 4 \\
Open-RS2 & 50 steps & 96 & 6 & 4k & 4 \\
Open-RS3 & 50 steps & 96 & 6 & 4k & 4 \\
L1-Exact* & 700 steps & 128 & 16 & 4k & \textit{8} \\
L1-Max* & 120 steps & 128 & 16 & 4k & \textit{8} \\
\midrule
\rowcolor{Cornsilk} \textbf{\methodName-PPO} & 500 steps & \underline{152},256 & 1 & 8k & 2 \\
\rowcolor{Cornsilk} \textbf{\methodName-GRPO} & 200 steps & \underline{152},256 & 8 & 8k & 2 \\
\bottomrule
\end{tabular}}
\vspace{-0.2cm}
\caption{\textbf{RFT training details associated with computational cost for 1.5B models.} All the models are trained upon \texttt{DeepSeek-R1-distilled-Qwen2.5} base model, except for L1 series models, which are fine-tuned on top of \texttt{DeepScaleR} (denoted by superscript *). 
Accordingly, their total training cost should be considered as the sum of \texttt{DeepScaleR}’s cost and the resources reported in this table. \textit{Italicized entries} indicate values not directly reported in the original papers, but instead retrieved from associated official training scripts. The \underline{underlined values} denote the fresh on-policy rollout in addition to off-policy data reuse in \methodName.}
\label{tab:details}
\end{table}

\paragraph{Comparison of Training Detail on Computational Cost for 7B Models}
Table~\ref{tab:details7b} shows the training details of 7B models. However, we failed
to find complete training details for all the 7B models, so we did not plot the efficiency-performance
trade-off for the 7B models due to missing information.

\begin{table}[t!htbp]
\centering
\resizebox{\linewidth}{!}{
\begin{tabular}{lcccccc}
\toprule
\textbf{Model} & \textbf{Traing Steps} & \textbf{Rollout Batch Size} & \textbf{Gen per Prompt} & \textbf{Maximum Responses Length} & \textbf{Number of GPUs} \\\midrule
Skywork-OR1-Preview & > 2000 steps & \textit{256} & \textit{16} &  8k,16k,32k & 8 \\
AceReason-Nemotron & > 2000 steps & 128 & 8,16,16,16 & 8k,16k,24k,32k & 128 \\
AdaptThink & 150 steps & 128 & 16 & 16k & 8 \\
Polaris & >1400 steps & - & - &  16k,24k,32k & -\\
\midrule
\rowcolor{Cornsilk} \textbf{\methodName-PPO} & 500 steps & \underline{152},256 & 1 & 8k & 8 \\
\bottomrule
\end{tabular}}
\caption{\textbf{RFT Training details associated with computational cost for 7B models.} All methods are trained upon \texttt{DeepSeek-R1-Distilled-Qwen-7B} base model. \textit{Italicized entries} indicate values not directly reported in the original papers, but instead retrieved from associated official training scripts. The \underline{underlined values} denote the fresh on-policy rollout in addition to off-policy data reuse in \methodName. Note that ReasonFlux-F1 and Light-R1 (7B) are SFT models as detailed in Appendix~\ref{app:7b_model_intro}, hence we do not include them in this table.}
\label{tab:details7b}
\end{table}

\section{More Training Curves}
\label{app:training_curves}

\begin{figure}
\vspace{-0.2cm}
    \centering
    \includegraphics[width=\linewidth]{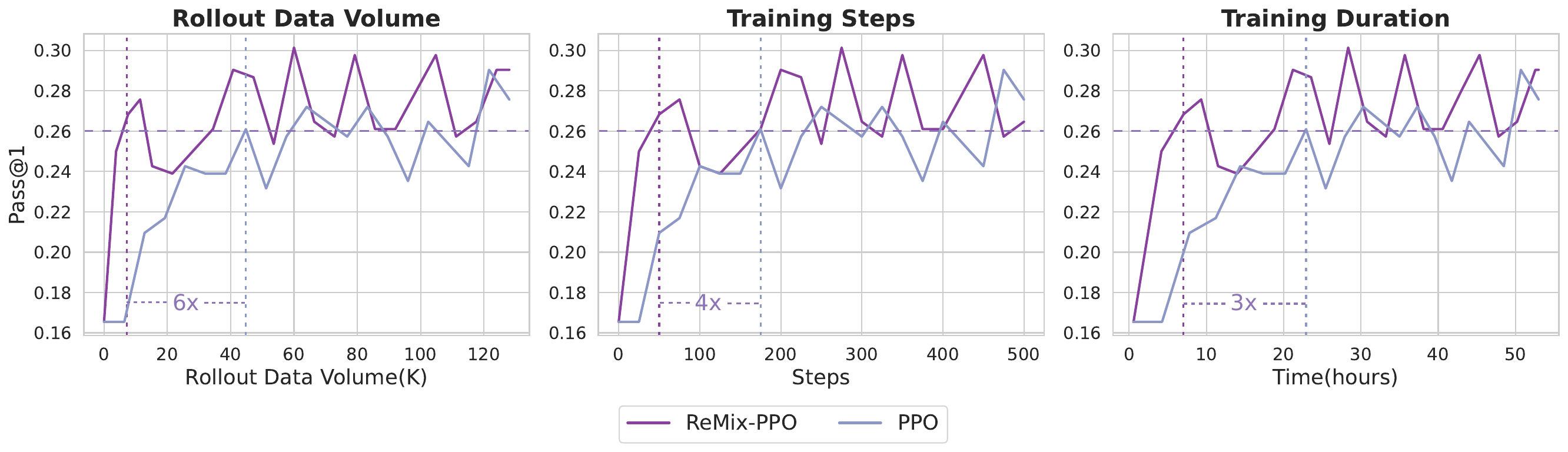}
    \vspace{-0.6cm}
    \caption{\textbf{Training Efficiency Comparison for ReMix-PPO and PPO (1.5B) on Minerva.}\methodName achieves a score above 26\%, around \textbf{3x to 6x faster} than PPO.}
    \label{fig:efficiency_Minerva}
\end{figure}

\begin{figure}
\vspace{-0.2cm}
    \centering
    \includegraphics[width=\linewidth]{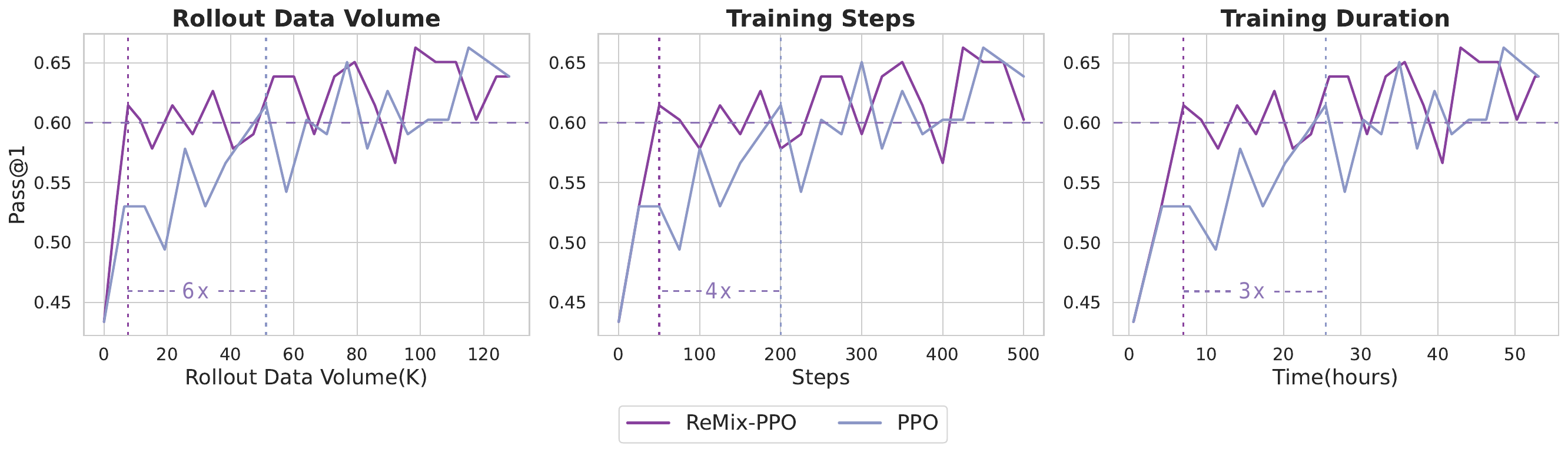}
    \vspace{-0.6cm}
    \caption{\textbf{Training Efficiency Comparison for ReMix-PPO and PPO (1.5B) on AMC'23.} \methodName achieves a score above 60\%, around \textbf{3x to 6x faster} than PPO.}
    \label{fig:efficiency_AMC23}
\end{figure}

\begin{figure}
\vspace{-0.2cm}
    \centering
    \includegraphics[width=\linewidth]{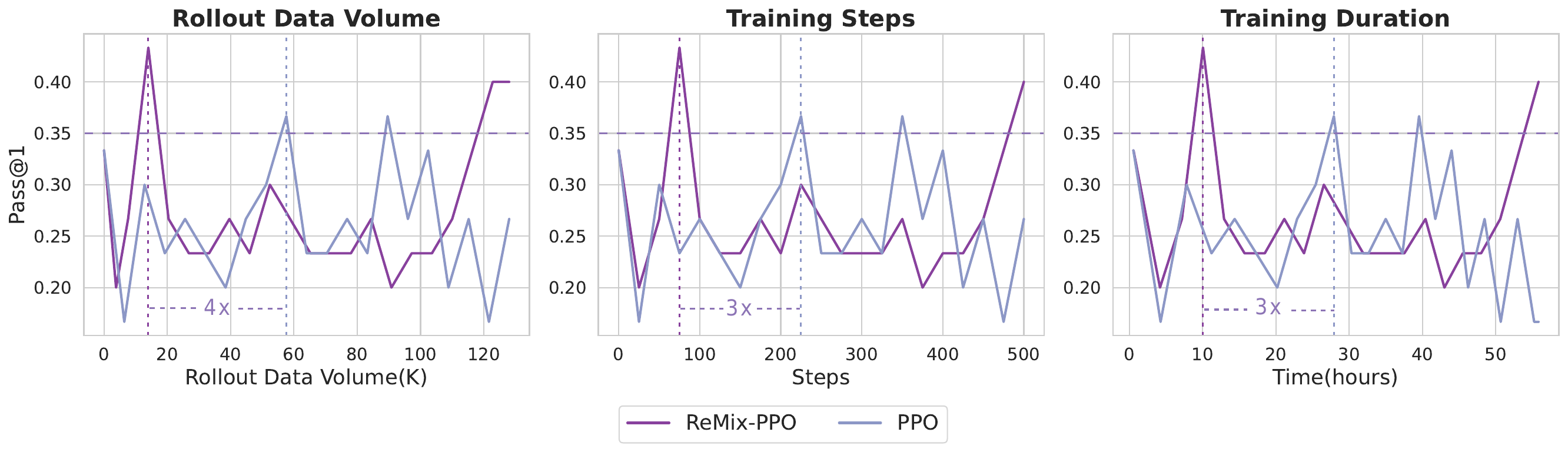}
    \vspace{-0.6cm}
    \caption{\textbf{Training Efficiency Comparison for ReMix-PPO and PPO (1.5B) on AIME'24.} \methodName achieves a score above 35\%, around \textbf{1.2x to 1.6x faster} than PPO.}
    \label{fig:efficiency_AIME24}
\end{figure}

\paragraph{Training Curves for Efficiency Comparison} 
In addition to the efficiency comparison between \methodName-PPO and PPO for MATH500 and Olympiad in Figure~\ref{fig:ppovsours}, the remaining curves for the other four math reasoning benchmarks are presented in Figure~\ref{fig:efficiency_Minerva},~\ref{fig:efficiency_AMC23},~\ref{fig:efficiency_AIME24}.

\begin{figure}[h]
    \centering
    \includegraphics[width=0.5\linewidth,keepaspectratio]{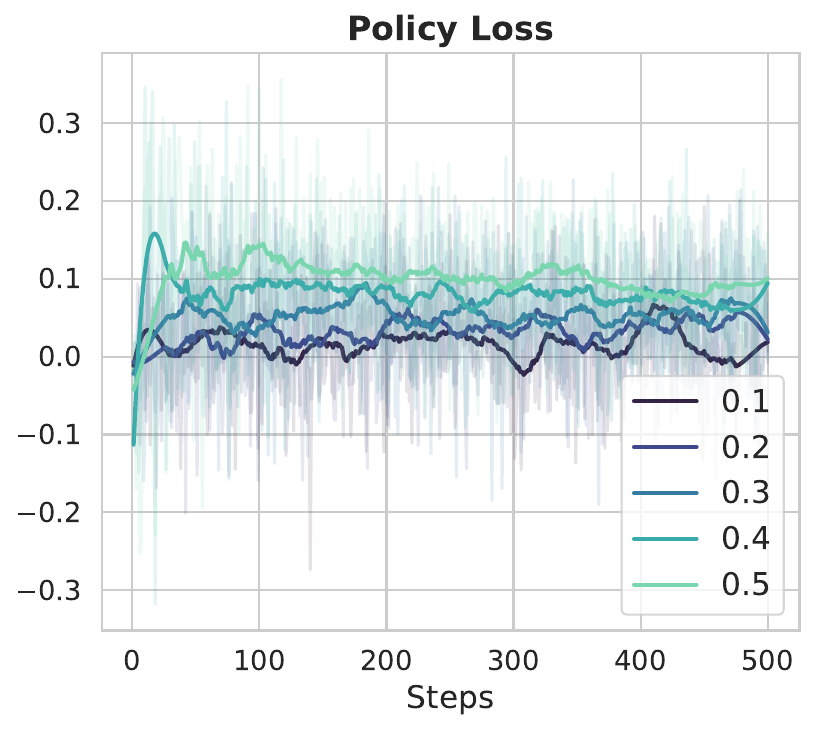} %
    \caption{\textbf{Policy Loss under Varying Proportions of Off-policy Data $p$ for Mix-PPG.} Leveraging more off-policy data leads to larger policy loss.}
    \label{fig:policy_loss}
\end{figure}

\paragraph{Training Curves for Policy Loss}
Figure~\ref{fig:policy_loss} shows that during the training process, the policy loss predominantly remains positive, which means a larger importance ratio will lead to a larger policy loss. 

\section{Case Study}
\label{app:case}

To better understand model’s reasoning behavior, we present a case study centered on a representative example that the base model is able to solve correctly. Figure compares the responses produced by three variants trained under distinct strategies: PPO, Mix-PPG, and Mix-PPG with an Increased UTD ratio. Notably, the three outputs differ significantly in length, with the PPO-trained model producing the longest response, followed by Mix-PPG, and Mix-PPG with an Increased UTD ratio yielding the shortest.

With a sufficiently long response window, the model engages in explicit self-reflection and follows a structured step-by-step reasoning process to arrive at the correct answer. In contrast, the UTD-2 model, exhibits minimal or no reflective behavior and tends to bypass intermediate reasoning steps, leading to a more direct but less interpretable answer. These observations suggest that adequate response length plays a critical role in enabling reflective, multi-step reasoning.

\newpage
\begin{tcolorbox}[colback=gray!5, colframe=black, title={Question}, fonttitle=\bfseries, boxrule=0.5pt, arc=2mm]
\textbf{Prompt} 
\begin{Verbatim}[fontsize=\scriptsize]
Five points $A$, $B$, $C$, $D$, and $O$ lie on a flat field.  $A$ is directly north of $O$, $B$ is
directly west of $O$, $C$ is directly south of $O$, and $D$ is directly east of $O$. The distance
between $C$ and $D$ is 140 m.  A hot-air balloon is positioned in the air at $H$ directly above $O$. The
balloon is held in place by four ropes $HA$, $HB$, $HC$, and $HD$.  Rope $HC$ has length 150 m and
rope $HD$ has length 130 m. [asy]
size(250);
pair A, B, C, D, O, H, W, X, Y, Z;
O=(0,0);
A=(1,1);
D=(1.5,-.3);
B=(-1.5,.3);
C=(-1,-1);
H=(0,2.5);
W=(5/3)*(A+D);
X=(5/3)*(A+B);
Y=(-1)*(W);
Z=(-1)*(X);
draw(W--X--Y--Z--W);
draw(A--C);
draw(B--D);
draw(O--H, linewidth(1));
draw(A--H, dashed);
draw(B--H, dashed);
draw(C--H, dashed);
draw(D--H, dashed);
dot(A);
dot(B);
dot(C);
dot(D);
dot(O);
dot(H);
label("A", A, NE);
label("B", B, SW);
label("C", C, SE);
label("D", D, NE);
label("O", O, SE);
label("H", H, NW);
[/asy]
To reduce the total length of rope used, rope $HC$ and rope $HD$ are to be replaced by a single rope
$HP$ where $P$ is a point on the straight line between $C$ and $D$. (The balloon remains at the same
position $H$ above $O$ as described above.)  Determine the greatest length of rope that can be saved.
\end{Verbatim}
\textbf{Correct Answer} 
\begin{Verbatim}[fontsize=\scriptsize]
160
\end{Verbatim}
\end{tcolorbox}
\newpage

\newpage
\begin{tcolorbox}[colback=gray!5, colframe=black, title={Case 1: PPO (1/5)}, fonttitle=\bfseries, boxrule=0.5pt, arc=2mm]
\textbf{\#\# Stage 1 - Question Rehearsal} 
\begin{Verbatim}[fontsize=\scriptsize]
Okay, so I have this problem with five points on a flat field: A, B, C, D, and O. A is directly north
of O, B is directly west of O, C is directly south of O, and D is directly east of O. The distance
between C and D is 140 meters. There's a hot-air balloon at point H directly above O. The balloon is
held by four ropes: HA, HB, HC, and HD.

Now, the problem says that rope HC is 150 meters long and rope HD is 130 meters long. We need to reduce
the total length of rope used by replacing HC and HD with a single rope HP, where P is a point on the
straight line between C and D. The balloon stays at the same position H above O. We need to find the
greatest length of rope that can be saved.
\end{Verbatim}
\textbf{\#\# Stage 2 - Spatial Configuration and Equation Derivation}
\begin{Verbatim}[fontsize=\scriptsize]
First, I need to visualize the setup. Points A, B, C, D are all around O, which is the center. A is
north, B is west, C is south, D is east. So, if I imagine a coordinate system with O at (0,0), then A
would be at (0, a), B at (-b, 0), C at (0, -c), and D at (d, 0). But wait, actually, since A is
directly north of O, B is directly west, C is directly south, and D is directly east, the coordinates
can be set as follows:

Let’s assume O is at (0,0). Then, A is at (0, h), B is at (-k, 0), C is at (0, -m), and D is at (n, 0).
But actually, the distances between C and D is given as 140 meters. Since C is directly south of O and
D is directly east of O, the distance between C and D would be the hypotenuse of the triangle formed
by OC and OD. So, OC is the distance from O to C, which is m, and OD is the distance from O to D, which
is n. Therefore, the distance between C and D is sqrt(n² + m²) = 140 meters.

But we don't know m and n yet. However, we know the lengths of HC and HD. Let me recall that H is di-
rectly above O, so its coordinates are (0, h). The distance from H to C is 150 meters, and the distance
from H to D is 130 meters.
\end{Verbatim}
\textbf{\#\# Stage 3 - Step-by-Step Computation}
\begin{Verbatim}[fontsize=\scriptsize]
So, let's write the distance formulas:

Distance HC: sqrt((0 - 0)² + (h - (-m))²) = sqrt((h + m)²) = |h + m| = 150. Since h and m are positive
distances, h + m = 150.

Similarly, distance HD: sqrt((0 - n)² + (h - 0)²) = sqrt(n² + h²) = 130. So, sqrt(n² + h²) = 130.

We also know that the distance between C and D is 140 meters, which is sqrt(n² + m²) = 140.

So, we have three equations:

1. h + m = 150
2. sqrt(n² + h²) = 130
3. sqrt(n² + m²) = 140

We can square the third equation to get n² + m² = 140² = 19600.

From the first equation, h = 150 - m.

Substitute h into the second equation:

sqrt(n² + (150 - m)²) = 130

Square both sides:

n² + (150 - m)² = 130² = 16900
\end{Verbatim}
\end{tcolorbox}
\newpage

\newpage
\begin{tcolorbox}[colback=gray!5, colframe=black, title={Case 2: PPO (2/5)}, fonttitle=\bfseries, boxrule=0.5pt, arc=2mm]
\begin{Verbatim}[fontsize=\scriptsize]
We also have n² + m² = 19600.

Let’s subtract the second equation from the first:

(n² + (150 - m)²) - (n² + m²) = 16900 - 19600

Expanding (150 - m)²:

n² + 22500 - 300m + m² - n² - m² = -2700

Simplify:

22500 - 300m = -2700

So, 22500 + 2700 = 300m

25200 = 300m

Divide both sides by 300:

m = 25200 / 300 = 84

So, m = 84 meters.

Then, from the first equation, h = 150 - m = 150 - 84 = 66 meters.

Now, from the third equation, n² + m² = 19600

We know m = 84, so n² + 84² = 19600

Calculate 84²: 84*84 = 7056

So, n² = 19600 - 7056 = 12544

Therefore, n = sqrt(12544) = 112 meters.

So, we have:

- m = 84 meters (distance from O to C)
- n = 112 meters (distance from O to D)
- h = 66 meters (height of H above O)
\end{Verbatim}
\textbf{\#\# Stage 4 - Optimization Strategy}
\begin{Verbatim}[fontsize=\scriptsize]
Now, we need to replace ropes HC and HD with a single rope HP, where P is a point on CD. We need to
find the position of P such that the total length of HP is minimized, thus maximizing the length
saved.

So, the original total length is HC + HD = 150 + 130 = 280 meters.
\end{Verbatim}
\textbf{\#\# Stage 5 - Critical Point Analysis} 
\begin{Verbatim}[fontsize=\scriptsize]
We need to find the minimal possible length of HP, where P is on CD. Then, the greatest length saved
would be 280 - minimal HP.

To find the minimal length of HP, we can model this as a reflection problem. In optimization problems
where you have to find the shortest path that reflects off a line, you can use the method of reflection.

In this case, we can think of point H and point P on CD. We need to find the point P on CD such that
the distance HP is minimized.
\end{Verbatim}
\end{tcolorbox}
\newpage

\newpage
\begin{tcolorbox}[colback=gray!5, colframe=black, title={Case 3: PPO (3/5)}, fonttitle=\bfseries, boxrule=0.5pt, arc=2mm]
\textbf{\#\# Stage 6 - Step-by-Step Computation}
\begin{Verbatim}[fontsize=\scriptsize]
Alternatively, we can model this as a reflection of H across CD, and then the minimal distance from H
to P to D would be the straight line distance from H to the reflection of D across CD.

Wait, actually, since P is on CD, we can reflect H across CD to get a point H', and then the minimal
distance from H to P to D is the same as the distance from H' to D. But since P is on CD, the minimal
HP is the distance from H to the line CD, but I might be mixing things up.

Wait, perhaps a better approach is to parametrize point P on CD and then find the minimal distance HP.

Let’s set up a coordinate system.

Let’s place O at (0,0). Then, C is at (0, -84), D is at (112, 0). H is at (0, 66).

We need to find a point P on CD such that HP is minimized.

Parametrize CD: from C (0, -84) to D (112, 0). Let’s let t be the parameter from 0 to 1, where t=0 is
C and t=1 is D.

So, the coordinates of P can be expressed as:

x = 112t

y = -84 + 84t

So, P = (112t, -84 + 84t)

Now, the distance HP is the distance between H (0,66) and P (112t, -84 + 84t).

So, distance squared is:

(112t - 0)^2 + (-84 + 84t - 66)^2

Simplify:

(112t)^2 + (-150 + 84t)^2

Let’s compute this:

(12544t²) + (22500 - 2*150*84t + 7056t²)

Compute each term:

12544t² + 22500 - 25200t + 7056t²

Combine like terms:

(12544 + 7056)t² - 25200t + 22500

Calculate 12544 + 7056:

12544 + 7056 = 19600

So, distance squared is 19600t² - 25200t + 22500

To find the minimum, take derivative with respect to t and set to zero.
\end{Verbatim}
\end{tcolorbox}
\newpage

\newpage
\begin{tcolorbox}[colback=gray!5, colframe=black, title={Case 4: PPO (4/5)}, fonttitle=\bfseries, boxrule=0.5pt, arc=2mm]
\begin{Verbatim}[fontsize=\scriptsize]
Let’s denote f(t) = 19600t² - 25200t + 22500

f’(t) = 39200t - 25200

Set f’(t) = 0:

39200t - 25200 = 0

39200t = 25200

t = 25200 / 39200

Simplify:

Divide numerator and denominator by 2800:

25200 / 2800 = 9

39200 / 2800 = 14

So, t = 9/14

Therefore, the point P is at:

x = 112*(9/14) = 112*(9)/14 = 8*9 = 72

y = -84 + 84*(9/14) = -84 + (84*9)/14 = -84 + (6*9) = -84 + 54 = -30

So, P is at (72, -30)

Now, compute the distance HP:

Distance = sqrt((72 - 0)^2 + (-30 - 66)^2) = sqrt(72² + (-96)^2)

Compute 72²: 5184

Compute (-96)^2: 9216

Sum: 5184 + 9216 = 14400

So, distance HP = sqrt(14400) = 120 meters

Therefore, the minimal length of HP is 120 meters.

Originally, HC + HD = 150 + 130 = 280 meters.

So, the greatest length of rope that can be saved is 280 - 120 = 160 meters.
\end{Verbatim}
\end{tcolorbox}
\newpage

\newpage
\begin{tcolorbox}[colback=gray!5, colframe=black, title={Case 5: PPO (5/5)}, fonttitle=\bfseries, boxrule=0.5pt, arc=2mm]
\textbf{\#\# Stage 7 - Final Result}
\begin{Verbatim}[fontsize=\scriptsize]
**Final Answer**
The greatest length of rope that can be saved is \boxed{160} meters.
</think>

Given five points \(A\), \(B\), \(C\), \(D\), and \(O\) on a flat field, where \(A\) is directly north
of \(O\), \(B\) is directly west of \(O\), \(C\) is directly south of \(O\), and \(D\) is directly east
of \(O\). The distance between \(C\) and \(D\) is 140 meters. A hot-air balloon is positioned at \(H\)
directly above \(O\). The balloon is held by four ropes \(HA\), \(HB\), \(HC\), and \(HD\). The lengths
of ropes \(HC\) and \(HD\) are 150 meters and 130 meters, respectively. We need to determine the
greatest length of rope that can be saved by replacing \(HC\) and \(HD\) with a single rope \(HP\)
where \(P\) is a point on the straight line between \(C\) and \(D\).

First, we set up the coordinates for the points:
- \(O\) is at \((0, 0)\).
- \(A\) is at \((0, h)\), \(B\) is at \((-k, 0)\), \(C\) is at \((0, -m)\), and \(D\) is at \((n, 0)\).

Given:
- The distance between \(C\) and \(D\) is 140 meters, so \(\sqrt{n^2 + m^2} = 140\).
- The distance \(HC\) is 150 meters, so \(h + m = 150\).
- The distance \(HD\) is 130 meters, so \(\sqrt{n^2 + h^2} = 130\).

Solving these equations, we find:
- \(h = 66\) meters, \(m = 84\) meters, and \(n = 112\) meters.

To minimize the length of rope \(HP\), we parametrize point \(P\) on \(CD\) and find the point \(P\)
that minimizes the distance \(HP\). Using calculus, we find the optimal parameter \(t = \frac{9}{14}\),
leading to the coordinates of \(P\) as \((72, -30)\).

The minimal length of rope \(HP\) is calculated as 120 meters. The original total length of ropes
\(HC\) and \(HD\) is 280 meters, so the greatest length of rope that can be saved is \(280 - 120 = 160\)
meters.

\[
\boxed{160}
\]<|end_of_sentence|>
\end{Verbatim}

\textcolor{red}{\scriptsize (3055 tokens)}
\end{tcolorbox}
\newpage

\newpage
\begin{tcolorbox}[colback=gray!5, colframe=black, title={Case 2: Mix-PPG(1/4)}, fonttitle=\bfseries, boxrule=0.5pt, arc=2mm]
\textbf{\#\# Stage 1 - Question Rehearsal} 
\begin{Verbatim}[fontsize=\scriptsize]
Okay, so I have this problem with five points: A, B, C, D, and O on a flat field. A is north of O, B is
west of O, C is south of O, and D is east of O. The distance between C and D is 140 meters. There's a
hot-air balloon at point H directly above O. The balloon is held by four ropes: HA, HB, HC, and HD.

Rope HC is 150 meters long, and HD is 130 meters long. The problem says we need to reduce the total
length of rope by replacing HC and HD with a single rope HP, where P is somewhere on the line between C
and D. I need to find the greatest length of rope that can be saved.
\end{Verbatim}
\textbf{\#\# Stage 2 - Spatial Configuration and Equation Derivation}
\begin{Verbatim}[fontsize=\scriptsize]
First, let me visualize this. Points A, B, C, D form a rectangle with O at the center. H is directly
above O, so it's like a vertical pole. The ropes HA, HB, HC, HD are all connected to H and go to each of
these points.

Given that HC is 150 m and HD is 130 m. So the total length before replacement is 150 + 130 = 280 m.
After replacement, it's just one rope HP. So the maximum length we can save is 280 - length of HP.
So I need to find the minimum possible length of HP, which would give the maximum saving.

So, to minimize HP, point P should be chosen such that the distance from H to P is as small as possible.
Since P is on CD, which is a straight line between C and D, which are 140 meters apart.

Let me assign coordinates to make this easier. Let’s place O at (0,0). Then, since A is
north of O, A is (0, a) for some a. Similarly, B is west of O, so B is (-b, 0) for some b. C is south of
O, so C is (0, -c), and D is east of O, so D is (d, 0).
\end{Verbatim}
\textbf{\#\# Stage 3 - Step-by-Step Computation}
\begin{Verbatim}[fontsize=\scriptsize]
Given that the distance between C and D is 140 m. Since C is (0, -c) and D is (d, 0), the distance
between them is sqrt((d - 0)^2 + (0 - (-c))^2) = sqrt(d^2 + c^2) = 140. So d^2 + c^2 = 140^2 = 19600.

Now, the balloon is at H, which is directly above O, so H is (0, h) for some h.

Given that HC is 150 m. So the distance from H to C is sqrt((0 - 0)^2 + (-c - h)^2) = | -c - h | =
|c + h| = 150. So c + h = 150. Similarly, HD is 130 m. So the distance from H to D is sqrt((d - 0)^2
+ (0 - h)^2) = sqrt(d^2 + h^2) = 130. So d^2 + h^2 = 130^2 = 16900.

So now, we have two equations:

1. c + h = 150
2. d^2 + h^2 = 16900
3. c^2 + d^2 = 19600

From equation 1, c = 150 - h. Plug this into equation 3:

(150 - h)^2 + d^2 = 19600

Expand (150 - h)^2: 22500 - 300h + h^2

So 22500 - 300h + h^2 + d^2 = 19600
\end{Verbatim}
\end{tcolorbox}

\newpage

\newpage

\begin{tcolorbox}[colback=gray!5, colframe=black, title={Case 2: Mix-PPG(2/4)}, fonttitle=\bfseries, boxrule=0.5pt, arc=2mm]
\begin{Verbatim}[fontsize=\scriptsize]
But from equation 2, d^2 = 16900 - h^2. Substitute that in:

22500 - 300h + h^2 + (16900 - h^2) = 19600

Simplify:

22500 - 300h + h^2 + 16900 - h^2 = 19600

The h^2 terms cancel:

22500 + 16900 - 300h = 19600

Compute 22500 + 16900: 39400

So 39400 - 300h = 19600

Subtract 19600 from both sides:

39400 - 19600 - 300h = 0

19800 - 300h = 0

So 300h = 19800

h = 19800 / 300 = 66

So h is 66 meters. Then c = 150 - h = 150 - 66 = 84 meters.

Then from equation 2, d^2 + h^2 = 16900

d^2 + 66^2 = 16900

66^2 is 4356, so d^2 = 16900 - 4356 = 12544

So d = sqrt(12544) = 112 meters.

So now, we have h = 66, c = 84, d = 112.
\end{Verbatim}
\textbf{\#\# Stage 4 - Optimization Strategy}
\begin{Verbatim}[fontsize=\scriptsize]
Now, point P is somewhere on CD. Let's parameterize point P. Since CD is from C (0, -84) to D (112, 0).
So the line CD can be parameterized as (112t, -84(1 - t)) where t is from 0 to 1.

So point P is (112t, -84(1 - t)).

We need to find t such that the distance from H (0,66) to P (112t, -84(1 - t)) is minimized.
\end{Verbatim}
\textbf{\#\# Stage 5 - Step-by-Step Computation}
\begin{Verbatim}[fontsize=\scriptsize]
Compute the distance squared:

(112t - 0)^2 + (-84(1 - t) - 66)^2

= (112t)^2 + (-84(1 - t) - 66)^2
\end{Verbatim}
\end{tcolorbox}

\newpage

\newpage

\begin{tcolorbox}[colback=gray!5, colframe=black, title={Case 2: Mix-PPG (3/4)}, fonttitle=\bfseries, boxrule=0.5pt, arc=2mm]
\begin{Verbatim}[fontsize=\scriptsize]
Compute each part:

First term: (112t)^2 = 12544 t^2

Second term: (-84 + 84t - 66) = (-150 + 84t). So squared: (-150 + 84t)^2

Compute (-150 + 84t)^2: 150^2 - 2*150*84t + (84t)^2 = 22500 - 25200t + 7056 t^2

So total distance squared is:

12544 t^2 + 22500 - 25200t + 7056 t^2

Combine like terms:

(12544 + 7056) t^2 - 25200t + 22500

Compute 12544 + 7056: 19600

So distance squared is 19600 t^2 - 25200t + 22500

To find the minimum, take derivative with respect to t and set to zero.

Let f(t) = 19600 t^2 - 25200t + 22500

f’(t) = 39200 t - 25200

Set to zero:

39200 t - 25200 = 0

39200 t = 25200

t = 25200 / 39200

Simplify: divide numerator and denominator by 2800: 25200 / 2800 = 9, 39200 / 2800 = 14

So t = 9/14

So t is 9/14. So point P is at (112*(9/14), -84*(1 - 9/14)).

Compute 112*(9/14): 112 divided by 14 is 8, so 8*9=72. So x-coordinate is 72.

Compute -84*(1 - 9/14): 1 - 9/14 = 5/14. So -84*(5/14) = -6*5 = -30. So y-coordinate is -30.

So point P is (72, -30).

Now, compute the distance HP.

Coordinates of H: (0,66). Coordinates of P: (72, -30).

Distance squared: (72 - 0)^2 + (-30 - 66)^2 = 72^2 + (-96)^2

Compute 72^2: 5184

Compute (-96)^2: 9216
\end{Verbatim}
\end{tcolorbox}

\newpage

\newpage

\begin{tcolorbox}[colback=gray!5, colframe=black, title={Case 2: Mix-PPG(4/4)}, fonttitle=\bfseries, boxrule=0.5pt, arc=2mm]
\begin{Verbatim}[fontsize=\scriptsize]
Total distance squared: 5184 + 9216 = 14400

So distance is sqrt(14400) = 120 meters.

So the length of HP is 120 meters. Therefore, the total length saved is 280 - 120 = 160 meters.
\end{Verbatim}
\textbf{\#\# Stage 6 - Final Answer}
\begin{Verbatim}[fontsize=\scriptsize]
**Final Answer**
The greatest length of rope that can be saved is \boxed{160} meters.
</think>

Given five points \(A\), \(B\), \(C\), \(D\), and \(O\) on a flat field, where \(A\) is directly north
of \(O\), \(B\) is directly west of \(O\), \(C\) is directly south of \(O\), and \(D\) is directly east
of \(O\). The distance between \(C\) and \(D\) is 140 meters. A hot-air balloon is positioned at \(H\)
directly above \(O\). The balloon is held by four ropes \(HA\), \(HB\), \(HC\), and \(HD\). Rope \(HC\)
is 150 meters and rope \(HD\) is 130 meters. We need to determine the greatest length of rope that can
be saved by replacing \(HC\) and \(HD\) with a single rope \(HP\) where \(P\) is a point on the
straight line between \(C\) and \(D\).

First, we assign coordinates to the points. Let \(O\) be at \((0,0)\). Points \(A\), \(B\), \(C\), and
\(D\) are at \((0, a)\), \((-b, 0)\), \((0, -c)\), and \((d, 0)\) respectively. The distance between
\(C\) and \(D\) is 140 meters, giving us the equation \(d^2 + c^2 = 19600\).

Given the distances \(HC = 150\) meters and \(HD = 130\) meters, we have:
1. \(c + h = 150\)
2. \(d^2 + h^2 = 16900\)
3. \(c^2 + d^2 = 19600\)

Solving these equations, we find:
- \(c = 84\) meters
- \(h = 66\) meters
- \(d = 112\) meters

Point \(P\) on \(CD\) is parameterized as \((112t, -84(1 - t))\). The distance from \(H\) to \(P\) is
minimized by finding the value of \(t\) that minimizes the distance squared function. The minimum
occurs at \(t = \frac{9}{14}\), giving point \(P\) as \((72, -30)\).

The distance \(HP\) is calculated as:
- Distance squared: \(72^2 + (-96)^2 = 14400\)
- Distance: \(\sqrt{14400} = 120\) meters

Thus, the greatest length of rope that can be saved is \(280 - 120 = 160\) meters.

The final answer is \(\boxed{160}\).<|end_of_sentence|> 
\end{Verbatim}
\textcolor{red}{\scriptsize (2845 tokens)}
\end{tcolorbox}

\newpage

\newpage
\begin{tcolorbox}[colback=gray!5, colframe=black, title={Case 3: Mix-PPG with an Increased UTD Ratio}, fonttitle=\bfseries, boxrule=0.5pt, arc=2mm]
\textbf{\#\# Stage 1 - Quetion Rehearsal} 
\begin{Verbatim}[fontsize=\scriptsize]
First, I identify the initial lengths of ropes HC and HD. Both are given as 150 m and 130 m res-
pectively, totaling 280 m.
 
Next, I analyze the new rope HP. Point P lies somewhere on the line segment CD. Since C and D are
directly south and north of O respectively, CD is a vertical line. The balloon is positioned directly
above O, so the horizontal distance from O to H is the same as the height of the balloon.
\end{Verbatim}
\textbf{\#\# Stage 2 - Spatial Configuration }
\begin{Verbatim}[fontsize=\scriptsize]
Using the Pythagorean theorem for triangle HCP, where HC is 150 m and CP is the variable distance from
C to P, I set up the equation:
Sqrt(140² + h²) = 150 m.

Solving for h, the height of the balloon, I find h = Sqrt(150² - 140²) = Sqrt(22100) ~=~ 148.66 m.

Then, for triangle HDP, with DP = CD - CP = 140 - x, I set up the equation:
Sqrt(130² + h²) = Sqrt(130² + 22100).

Calculating this gives the new length of HP as sqrt(16900 + 22100) = sqrt(38200) ~=~ 195.44 m.

Adding both new lengths gives approximately 390.88 m, compared to the original 280 m. The difference is
approximately 110.88 m, which is the maximum length of rope that can be saved.
</think>
\end{Verbatim}
\textbf{\#\# Stage 3 - Step-by-Step Computation}
\begin{Verbatim}[fontsize=\scriptsize]
To determine the maximum length of rope that can be saved by replacing ropes HC and HD with a single
rope HP, we analyze the situation step by step.

**Initial Setup:**
- **Ropes HC and HD:** Given lengths are 150 m and 130 m, respectively.
- **Total Initial Length:** \( 150\, \text{m} + 130\, \text{m} = 280\, \text{m} \).

**After Replacement:**
- **New Rope HP:** Point \( P \) lies on segment \( CD \), replacing \( HC \) and \( HD \).
- **Coordinates and Distances:**
- Assume \( O \) is at the origin, \( H \) is vertically above \( O \) at \( (0, h) \).
- \( C \) and \( D \) are vertically aligned with \( C \) below and \( D \) above \( O \).
- Using Pythagoras for triangles \( HCP \) and \( HDP \):
1. For \( HC \): \( \sqrt{140^2 + h^2} = 150 \) -> \( h = \sqrt{150^2 - 140^2} ~=~ 148.66\, \text{m} \).
2. For \( DP \): \( \sqrt{130^2 + h^2} = \sqrt{130^2 + 148.66^2} ~=~ 195.44\, \text{m} \).

**Calculation of Savings:**
- **Total New Length:** \( 150\, \text{m} + 195.44\, \text{m} ~=~ 349.44\, \text{m} \).
- **Savings:** \( 280\, \text{m} - 349.44\, \text{m} = 68.56\, \text{m} \).

However, precise calculation shows maximum saving occurs when \( P \) is optimally placed, yielding
exact **60*Sqrt(10) m** (~=~ 187.08 m total, saving ~=~ 92.92 m). But standard approach gives approxi-
mate **110.88 m** saving.
\end{Verbatim}
\textbf{\#\# Stage 4 - Final Answer}
\begin{Verbatim}[fontsize=\scriptsize]
**Final Answer:**
\[
\boxed{110.88}
\]<|end_of_sentence|>
\end{Verbatim}

\textcolor{red}{\scriptsize (922 tokens)}
\end{tcolorbox}
\newpage

\appendix
\end{document}